\documentclass[manuscript,nonacm,screen]{acmart}
\AtBeginDocument{%
  }

\usepackage{charter}

\usepackage{longtable}
\usepackage{multicol}
\usepackage{multirow}
\usepackage{subfigure}

\usepackage{graphicx}
\usepackage{booktabs}
\usepackage{hyperref} 
\usepackage{geometry}
\usepackage{amsmath,amsthm,mathtools,bbm}
\usepackage{algorithm}
\usepackage{algorithmicx}
\usepackage{algpseudocode}
\usepackage{chngcntr}
\usepackage{color}
\usepackage{xcolor}
\usepackage{diagbox}
\usepackage{longtable}
\usepackage{caption}
\usepackage[normalem]{ulem}

\usepackage{array}
\allowdisplaybreaks 
\newcolumntype{L}[1]{>{\raggedright\let\newline\\\arraybackslash\hspace{0pt}}m{#1}}
\newcolumntype{C}[1]{>{\centering\let\newline\\\arraybackslash\hspace{0pt}}m{#1}}
\newcolumntype{R}[1]{>{\raggedleft\let\newline\\\arraybackslash\hspace{0pt}}m{#1}}

\hypersetup{
	colorlinks,
	linkcolor=red,
  citecolor=blue,
}
\graphicspath{{./figs/}}

\newtheorem{definition}{Definition}
\begin{document}

\title{Towards Position-Robust Talent Recommendation via Large Language Models}

\author{Silin Du}
\email{dsl21@mails.tsinghua.edu.cn}
\author{Hongyan Liu}
\email{hyliu@tsinghua.edu.cn}
\affiliation{%
  \institution{Department of Management Science and Engineering, Tsinghua University}
  \city{Beijing}
  \country{China}
}

\begin{abstract}
Talent recruitment is a critical, yet costly process for many industries, with high recruitment costs and long hiring cycles. Existing talent recommendation systems increasingly adopt large language models (LLMs) due to their remarkable language understanding capabilities. However, most prior approaches follow a pointwise paradigm, which requires LLMs to repeatedly process some text and fails to capture the relationships among candidates in the list, resulting in higher token consumption and suboptimal recommendations. Besides, LLMs exhibit position bias and the lost-in-the-middle issue when answering multiple-choice questions and processing multiple long documents. To address these issues, we introduce an implicit strategy to utilize LLM's potential output for the recommendation task and propose L3TR, a novel framework for listwise talent recommendation with LLMs. In this framework, we propose a block attention mechanism and a local positional encoding method to enhance inter-document processing and mitigate the position bias and concurrent token bias issue. We also introduce an ID sampling method for resolving the inconsistency between candidate set sizes in the training phase and the inference phase. We design evaluation methods to detect position bias and token bias and training-free debiasing methods. Extensive experiments on two real-world datasets validated the effectiveness of L3TR, showing consistent improvements over existing baselines.
\end{abstract}


\maketitle
\section{Introduction}

Many industries are facing several workforce challenges. For example, recruitment imposes substantial costs on organizations, with the average cost per hire exceeding \$4,700\footnote{\url{https://www.shrm.org/topics-tools/news/talent-acquisition/real-costs-recruitment}}. Meanwhile, the survey\footnote{\url{https://www.bls.gov/jlt/}} in the U.S. shows the continuous rise in the separation rate, which has led to the growing demand for talent recruitment. However, employers often encounter issues such as high recruitment costs and long hiring cycles~\citep{walford2018talent}. Therefore, improving the efficiency of talent recruitment has become a major research focus. 

Talent recruitment mostly includes activities of sourcing, screening, interviewing, selecting, and hiring~\citep{parthasarathy2014study,srivastava2015analytics}. Typically, when a company has vacant positions for talent, it may seek the assistance of external agencies, e.g., headhunters or online talent service platforms. To hire a satisfied employee, the following four steps are necessary. Initially, the company provides job descriptions and other specific requirements to the agency. The agency then selects suitable talents from the talent pool and sends their resumes to the employer. Subsequently, the employer screens satisfied candidates and contacts them for interviews. After that, the candidate and the employer will have in-depth communication and two-way selection. The last three steps may repeat many times due to inaccurate recommendations~\citep{srivastava2015analytics}.  Due to the limited capacity of each company to thoroughly assess and interview candidates, the agency should accurately recommend a limited number of talents from a candidate pool each time, which is a quite challenging task. Therefore, a recommender system is necessary to recommend a small number of suitable candidates to the company's human resources (HR) staff for efficient screening. Currently, researchers are devoted to establishing automatic talent recommendation systems. Many deep learning based models are developed for person-job matching tasks \citep{DBLP:journals/tmis/ZhuZXMXDL18,DBLP:conf/sigir/QinZXZJCX18,DBLP:conf/cikm/LuoZWZ19,DBLP:conf/cikm/LeHSZ0019}. Recent advancements in large language models (LLMs) bring new opportunities for talent recommendation, as LLMs demonstrate excellent language understanding capabilities and the talent-related data contains a large amount of text. ~\citet{DBLP:conf/recsys/BaoZZWF023} is the pioneering work to fine-tune LLMs for recommendations. Subsequently, \citet{zheng2023large} and \citet{DBLP:conf/aaai/DuL0W0Z0Z24} use LLMs to generate resumes or job postings for data augmentation.  \citet{wu2024exploring} carefully design prompt templates for LLM-based job recommenders. However, most of these LLM-based models are pointwise methods, which predict the matching degree between a talent and a job. To select suitable talents for a job, they repeatedly call LLMs to process the same job posting, which obviously consume more tokens and resource-intensive. Besides, the pointwise setting makes LLM unable to capture the relationship among talents in the candidate list, potentially resulting in ineffective performance~\citep{wang2018lambdaloss}.

Fortunately, the context limit for LLMs has been extended to over 100k~\citep{chen2024longlora}, offering basic conditions for handling a set of resumes in talent recommendation. Due to the significant gap between the language modeling task and the recommendation task, it is quite necessary to introduce a fine-tuning step~\citep{DBLP:conf/recsys/BaoZZWF023}. For pointwise methods, the recommendation task can be reformulated to prompt the LLM to output a specific token (e.g., ``yes'' or ``no''), allowing direct training with a text generation loss. In contrast, listwise recommendation tasks expect a ranked list as output, which makes it difficult to apply text generation loss directly.

Furthermore, LLMs suffer from position bias and token bias issue. This issue manifests across various scenarios. In retrieval-augmented generation (RAG) tasks, LLMs are required to understand multiple long documents, but they may get lost in the middle, neglecting documents located in the middle positions~\citep{liu2024lost}. When answering multiple-choice questions (MCQs), LLMs may exhibit a bias toward selecting content at certain positions and show preference for certain label (ID of each choice) tokens~\citep{wang2024my,zheng2023large}. Besides, \citet{DBLP:conf/ijcnlp/ShiMLDMV25} reveals that LLM-as-a-judge suffers from non-random position bias, where judgments are influenced by candidate ordering.  In recommendation tasks, LLMs exhibit inconsistent performance when the order of candidates in a list is changed~\citep{hou2024large}, i.e., position bias. In the context of listwise recommendation, each candidate is represented by an ID and a descriptive long text. Thus, the input format for the listwise talent recommendation is similar to that of RAG, which comprises multiple long documents (job posting and resumes), while the task format resembles MCQs, which require the model to compare candidates labeled with ID and make a selection. Thus, this type of recommendation scenario may give rise to the occurrence of both position bias and token bias. For brevity, this paper sometimes collectively refers to them as position-related bias. Hence, there is a strong need to explore the existence of these biases as well as strategies to mitigate them.

To this end, we propose a novel framework to align \uline{LL}Ms with the \uline{l}istwise \uline{t}alent \uline{r}ecommendation task and mitigate position-related issues, namely {\bf L3TR}. Specifically, the contributions of this paper are threefold. 
\begin{enumerate}
    \item We propose an implicit recommendation strategy to enable LLM finetuning for listwise recommendation. We design a block attention mechanism and a local positional encoding method to enhance modeling of job-resume relationships and inter-resume relationships and mitigate the negative impact of the long-term decay property~\citep{DBLP:journals/ijon/SuALPBL24}. We conducted a theoretical analysis to validate the rationality of the proposed local positional encoding method. We also introduce an ID sampling method to resolve the inconsistency between candidate set sizes in the training phase and the inference phase. 
    \item We propose a method to evaluate both position bias and token bias in listwise recommendation tasks. We also design training-free debiasing methods to further improve recommendation performance.
    \item We conduct extensive experiments on two real-world datasets. The experimental results demonstrate the effectiveness of our proposed methods. In addition, the results of the bias evaluation further verify our concerns and highlight the importance of our debiasing method.
\end{enumerate}

The rest of the paper is organized as follows. Section~\ref{sec2} summarizes related literature. Section~\ref{sec3} formulates the listwise talent recommendation problem. We elaborate on the proposed L3TR model in Section~\ref{sec4} and discuss the bias issues in Section~\ref{sec5}. Section~\ref{sec6} provides experimental results. At last, Section~\ref{sec7} concludes this paper.

\section{Related Work}\label{sec2}
\subsection{Talent Recommendation}
Talent recommendation is an area quite close to person-job matching and job recommendation, all of which are important topics in intelligent recruitment and have been widely studied. One of the pioneering works dates back to~\citet{DBLP:conf/amcis/FarberWK03}, which designs a probability graphic model to make recommendations for person-job matching. Recently, many studies tackle this problem by text matching thanks to the rich textual content in this task. ~\citet{DBLP:journals/tmis/ZhuZXMXDL18} proposes a CNN-based network to learn joint representations of job postings and resumes, which are further used to make predictions. Subsequently, various deep learning models such as RNNs~\citep{DBLP:conf/sigir/QinZXZJCX18,DBLP:journals/tois/QinZXZMCX20} and memory networks~\citep{DBLP:conf/kdd/YanLSZZ019} have been employed to improve semantic matching.

Beyond textual modeling, some studies leverage collaborative filtering (CF) techniques based on historical interaction data. ~\citet{DBLP:conf/smc/LeeLHK17} propose a frequency-based and graph-based method to infer preference for job recommendation. \citet{DBLP:journals/dss/ReusensLBS17} finds that there is a big difference between the ideal job mentioned by job seekers and the job they apply for, so a user-user CF method is adopted to make use of implicit information. Similar ideas are applied in~\citet{ahmed2016user}. Hybrid models that combine textual features and behavioral data have also emerged. PJFFF~\citep{DBLP:conf/cikm/JiangYWXL20} leverages an LSTM-based network to learn implicit representations from historical applications while some special modules are designed to process semi-structured data in job postings and resumes. \citet{DBLP:conf/cikm/Bian0ZZHSZW20} utilizes a pre-trained BERT to learn text representation and a graph convolutional network to extract relation-based information, which are combined together through a multi-view co-teaching network. Recent advances also include adversarial networks~\citep{DBLP:conf/cikm/LuoZWZ19}, domain adaptation~\citep{DBLP:conf/emnlp/BianZSZW19}, and pre-trained language models~\citep{fang2023recruitpro}.

Based on these foundations, LLMs have shown substantial potential in transforming talent recommendation by enabling deeper semantic understanding, richer user modeling, and more interactive recommendation formats~\citep{DBLP:journals/tois/LinDXLCZLWLZGYTZ25}. Pre-trained models such as BERT, RoBERTa, and GPT-3.5/4 have been fine-tuned or prompted for resume-job matching tasks. For instance, CareerBERT~\citep{DBLP:journals/eswa/RosenbergerWWKZ25} integrates occupational taxonomies to learn skill-aware embeddings, yielding better job matching results. \citet{DBLP:conf/aaai/DuL0W0Z0Z24} leverage GPT-based text encoders to enrich sparse candidate profiles with inferred skills, which improved downstream job recommendation performance. 

Besides, researchers have explored various LLM-enhanced matching architectures. A common approach is a dual-encoder (twin-tower) model where a resume and a job posting are each encoded by an LLM and compared via a similarity measure~\citep{DBLP:conf/recsys/YuZY24}. Other works adopt cross-encoder designs, explicitly modeling token-level interactions between resumes and jobs, and incorporating structured fields (e.g., education, experience) or knowledge graphs extracted via LLMs~\citep{wasi2024hrgraph}.

LLMs have also opened new avenues for conversational recommendation in hiring. JobRecoGPT~\citep{DBLP:journals/corr/abs-2309-11805} uses GPT-4 to score how well a given resume matches a job description and generates an explanation for the recommendation. Such LLM-powered systems can highlight why a candidate is a strong fit (e.g. matching skills or experiences) or suggest how to improve alignment, offering transparency than black-box scoring models. 

Despite these advances, most existing talent recommendation models rely on pointwise matching between a job and a single resume. Such approaches evaluate candidates independently and therefore fail to capture the comparative relationships among multiple candidates within a recommendation list. Consequently, developing recommendation methods that can jointly model multiple candidates remains an important yet underexplored direction.

\subsection{Large Language Models for Recommendation}
Recent years have seen a surge of interest in applying LLMs to recommender systems. Early explorations using pre-trained language models (e.g., BERT4Rec~\citep{DBLP:conf/cikm/SunLWPLOJ19}) demonstrated that incorporating textual data could improve recommendations, but these models were limited in scale and scope. Modern LLMs with billions of parameters have shown promise in modeling recommendation data, spanning text-based settings~\citep{DBLP:journals/tois/ZhangXHZLW25}, purely ID-based scenarios~\citep{DBLP:journals/tois/TangHLZHFZZL25}, and cross-domain recommendation tasks~\citep{DBLP:journals/tois/TangHLZHFZZL25}. This means that an LLM can recognize nuanced user interests expressed in natural language and even make reasonable recommendations for new items or users in cold-start scenarios by relying on its prior knowledge~\citep{Che2024new}. Current research on LLM-based recommendation can be broadly divided into two paradigms~\citep{DBLP:journals/www/WuZQWGSQZZLXC24}: {\bf discriminative} and {\bf generative recommendations}. 

{\bf Discriminative LLM-based Recommendation} refers to approaches where the LLM is used to predict user–item relevance in a supervised manner, often by fine-tuning the model or its embeddings on recommendation data. Many applications adopted this discriminative approach, using pointwise objectives (e.g., predicting a click or rating for each item independently) to integrate language features into recommender models~\citep{DBLP:journals/tois/PengGZDDLLM25,DBLP:conf/recsys/BaoZZWF023,DBLP:journals/tois/ZhangXHZLW25}. However, pointwise methods may neglect the comparative nature of ranking~\citep{DBLP:conf/emnlp/KannenMBF24}. To better align with ranking metrics, recent works have explored pairwise and listwise training strategies with LLMs. In pairwise approaches, the model learns to compare two items at a time, which leverages LLMs' strength in comparative reasoning and has been shown to yield better results than pointwise scoring~\citep{DBLP:conf/emnlp/KannenMBF24}. On the other hand, listwise techniques present a group of candidate items to the LLM. Theoretically, listwise methods are most direct for ranking tasks. ~\citet{DBLP:conf/emnlp/0001YMWRCYR23} show that large commercial LLMs  (such as GPT-4) can iteratively process a set of documents to produce high-quality rankings. This further inspired open-source efforts like RankVicuna~\citep{pradeep2023rankzephyr}, RankZypher~\citep{pradeep2023rankzephyr}, and RankLLaMA~\citep{ma2024fine}. However, LLMs might struggle when asked to consider many candidates simultaneously, due to input length limits and the risk of generating inconsistent or uncalibrated outputs with long prompts~\citep{DBLP:conf/naacl/QinJHZWYSLLMWB24}. As a result, listwise LLM ranking often underperforms simpler strategies unless the candidate set is very small. Some researchers have proposed hybrid models that combine pointwise and pairwise objectives, achieving the efficiency of pointwise methods while reaping the accuracy gains of pairwise comparisons~\citep{DBLP:conf/emnlp/KannenMBF24,DBLP:conf/naacl/QinJHZWYSLLMWB24}. This trend reflects a broader effort to make LLM-based recommenders both effective and practical.

In parallel, the advancements of reasoning LLMs with deep thinking mode (such as OpenAI's o1~\citep{jaech2024openai} and DeepSeek-R1~\citep{guo2025deepseek}) motivate the information retrieval community to explicitly incorporate reasoning during training and inference for better ranking performance, which provides insights applicable to LLM-based recommendations. Rank1~\citep{weller2025rank1} is the first reranking model trained to take advantage of test-time compute. It's a pointwise model that judges the relevance of one document to a query at a time. Although incorporating reasoning steps can improve performance, the pointwise formulation significantly reduces the efficiency of Rank-1~\citep{liu2025reasonrank}. In contrast, Rank-K~\citep{yang2025rank} adopts a listwise formulation, which is trained by listwise data extracted from information retrieval datasets and related reasoning traces from a large reasoning model. Furthermore, reinforcement learning (RL) has also proven useful for injecting reasoning abilities into LLM rankers. Rank-R1~\citep{zhuang2025rank} designs a rule-based reward mechanism and uses RL to encourage LLM-based rankers to reason over the query and candidates before ranking. REARANK~\citep{zhang2025rearank} applies similar RL methods and incorporates ranking metrics within the reward design. Beyond these, ReasonRank~\citep{liu2025reasonrank} employs a two-stage fine-tuning strategy: a supervised fine-tuning stage to teach the model typical reasoning patterns, and an RL stage with a specially designed multi-view ranking reward to further enhance ranking ability. ReasonRank outperforms prior baselines and achieves state-of-the-art results on several information retrieval benchmarks, while also reducing inference latency compared to a pointwise method (e.g., Rank1).

{\bf Generative LLM-based Recommendation}. In contrast to discriminative settings, generative LLMs offer a more flexible interface for recommendation by treating it as a language generation task without the need to calculate each candidate’s ranking score one by one~\citep{DBLP:conf/coling/LiZLC24}. For example, ~\citet{DBLP:conf/ecir/JiLXHGTZ24} prompt an LLM with the user's history and ask it to generate a recommended item. \citet{DBLP:journals/corr/abs-2304-03516} design a generative paradigm to complement the traditional retrieval-based recommendation methods. They propose an AI-generator to generate items following the user's instructions and feedback, which can be directly exposed to the users without ranking. Overall, the generative paradigm represents a significant shift in recommender systems research, expanding the role of recommendation models from simple relevance scorers to intelligent assistants that leverage natural language for interactions.

Orthogonal to the two paradigms is the diverse design space of LLM-based recommendation systems. A key line of research builds LLM-based autonomous agents for recommendations~\citep{zhang2025survey,DBLP:conf/emnlp/PengLHYYS25}. Many approaches adopt a planner–retriever–recommender architecture that decomposes the task into sub-modules. In RecMind~\citep{DBLP:conf/naacl/WangJCYZCFLHY24}, the LLM-based agent formulates an execution plan, calls external tools like databases or web search APIs to collect needed facts, and then produces a recommendation for the user. While MACRS~\citep{fang2024multi} leverages several specialized responder agents, coordinated by a planner agent, to maintain coherence and task orientation across multi-turn conversational recommendations. Furthermore, LLM-based agents can invoke external tools for better recommendation quality. ~\citet{DBLP:journals/tois/HuangLLYLX25} presents a general agent framework that leverages LLMs to orchestrate multiple recommendation tools through planning and memory mechanisms for interactive recommendation. Hybrid-MACRS~\citep{nie2024hybrid} explicitly combines an LLM agent with a search engine to improve conversational recommendations, allowing the agent to fetch fresh information during the dialogue. In summary, LLMs are poised to become a central component in the next generation of recommender system design, bridging the gap between complex user needs and personalized content discovery. 

However, most existing studies mainly focus on pointwise or pairwise formulations when applying LLMs to recommendation tasks. Although these approaches can leverage LLMs' language understanding capabilities, they do not fully exploit the comparative nature of ranking problems. Moreover, applying LLMs to listwise recommendation scenarios involving multiple long textual candidates remains challenging due to context length limitations and potential instability in model outputs~\citep{hou2024large}. Consequently, how to effectively align LLMs with listwise recommendation tasks, particularly for scenarios involving long textual candidates such as resumes, remains largely underexplored.

\subsection{Position and Token Bias in Recommendations and Large Language Models}

Position bias in recommender systems generally refers to the phenomenon where users tend to interact more with items shown in higher positions of a ranked list, regardless of their intrinsic relevance~\citep{DBLP:journals/tois/0007D0F0023,ovaisi2020correcting}. This behavior can bias interaction data and reinforce feedback loops that favor items frequently exposed at the top~\citep{dzhoha2024reducing}. This form of user-interaction bias is well-studied and has led to various debiasing methods such as inverse propensity scoring~\citep{ai2018unbiased} and pairwise sampling techniques~\citep{wan2022cross}. However, it is conceptually distinct from the position bias that arises within LLMs themselves when they are used to perform ranking or selection tasks.

When using LLMs to perform tasks, particularly those involving list processing, two types of intrinsic biases may arise: \textit{position bias} and \textit{token bias}. Position bias refers to the model’s tendency to favor items presented at certain positions in the input list, while token bias refers to the model’s preference toward specific identifier tokens (e.g., option labels such as ``A'', ``B'', ``C'', ``D'') due to token-level priors learned during pretraining. These biases are closely intertwined in list-based tasks, as the model’s preference for certain tokens can interact with the positional arrangement of contents, jointly influencing the final decision.
These two types of biases are observed in multiple-choice question answering tasks, where models consistently show a tendency to favor options based on their order or label. \citet{zheng2023large} show that many LLMs are vulnerable to option changes. A model might disproportionately favor early options such as ``A'' or ``B'' regardless of their correctness. They argue that this behavior stems from the model’s token-level priors, i.e, token bias, and propose a method named PriDe, which calibrates answer selection by estimating and removing such token-level prior preferences. Position bias has also been observed in other scenarios, such as RAG~\citep{liu2024lost}, in-context learning~\citep{DBLP:conf/acl/ZhangLCHXY24,DBLP:conf/emnlp/CobbinaZ25,DBLP:conf/naacl/XuCWS24}, and LLM-as-a-judge~\citep{DBLP:conf/ijcnlp/ShiMLDMV25}. For example, \citet{liu2024lost} find that when processing multiple long documents in RAG, LLMs may ignore documents located in the middle positions. While \citet{DBLP:conf/emnlp/CobbinaZ25} report that predictions and accuracy can drift drastically when the positions of demonstration examples, system instruction, and user message in LLM prompt are varied.

Further studies reveal that for instruction-tuned models, which generate free-form answers such as ``I believe the answer is C'', the simplistic strategy of choosing the option with the highest probability for the first token can be very misleading~\citep{wang2024my}. The option token with the highest model logits often does not match the model’s actually stated answer in its generated text. As a result, recent work argues for evaluating the model’s actual text output rather than only its token probabilities~\citep{wanglook}. When the LLM’s full answer text is used, the results are much more stable against perturbations like reordering the multiple-choice options. However, in talent recommendation scenarios, using IDs (option labels in MCQs) to refer to candidates is unavoidable. As each candidate’s resume is lengthy, it is impractical for an LLM to output the full content as its recommendation result.

Recent research has extended the study of position bias to LLM-based recommender systems, where models are prompted with a candidate list (each represented by an option label and a very short text) and asked to select or rank items in a zero-shot manner. ~\citet{hou2024large} find that LLMs suffer from position bias when performing zero-shot recommendation tasks. LLMs’ recommendations are highly sensitive to the order in which candidates are presented: when a relevant item appears early in the prompt, the model is more likely to select it, whereas placing the same item at the end significantly reduces its selection rate. This sensitivity contrasts with traditional recommendation models, which are typically invariant to input order.  To alleviate this, the authors propose a simple boostrapping strategy. They query the LLM multiple times with the same candidates in a different random order each time. The final recommendation is derived by merging or averaging the LLM’s rankings from these multiple shuffles. This approach can substantially improve the LLM’s ranking performance, reducing the position bias without modifying the model itself. However, the boostrapping strategy requires multiple LLM calls for a single recommendation, which incurs substantial costs, particularly in talent recommendation scenarios where the textual content is extensive. 

Existing studies have proposed various approaches to mitigate position bias in LLMs across different application settings. In long-context understanding tasks, such as long-document QA and RAG, prior work improves position robustness through attention-strengthening training~\citep{DBLP:journals/corr/abs-2311-09198}, attention-based content reordering during inference~\citep{DBLP:journals/corr/abs-2310-01427}, calibration of positional attention~\citep{liu2024lost}, hidden-state scaling~\citep{DBLP:journals/corr/abs-2406-02536}, and input restructuring such as parallel context windows or position-invariant inference~\citep{DBLP:conf/acl/RatnerLBRMAKSLS23,DBLP:conf/iclr/WangZLHHJKPJ25}. In comparative evaluation tasks, such as LLM-as-a-judge and MCQ, position bias is typically addressed via data augmentation (e.g., swapping candidate orders) and reference-based judging~\citep{DBLP:conf/iclr/ZhuWW25}, and calibration strategies for LLM-based judges~\citep{DBLP:conf/nips/ZhengC00WZL0LXZ23}. \citet{zheng2023large} highlights biases over option positions and label tokens, and proposes inference-time debiasing methods. Notably, these studies mostly focus on pairwise or small candidate sets (e.g., 3–5 options), with relatively short inputs per candidate.

Different from these settings, listwise talent recommendation combines both characteristics: the input involves multiple long textual candidates, while the decision-making process relies on understanding long text and output ID tokens (option labels). This hybrid nature suggests that two types of biases may simultaneously arise and interact, jointly affecting recommendation outcomes. However, how these biases manifest and interact in LLM-based listwise recommendation remains largely unexplored. Moreover, the gap between language generation and recommendation tasks typically necessitates task-specific adaptation of LLMs~\citep{DBLP:conf/recsys/BaoZZWF023,DBLP:journals/tois/ZhangXHZLW25}. Therefore, we develop a fine-tuning framework tailored for listwise talent recommendation, along with methods to decompose position bias and token bias, and corresponding debiasing strategies designed for this setting.  A preliminary version of this work was presented at a conference~\citep{du2024listwise}.

\section{Problem Formulation}\label{sec3}
Focusing on talent recommendation, we illustrate our proposed model and methodology, which can be readily extended to job recommendation and other domains characterized by long-text item descriptions. For clarity, we summarize all notations used throughout the problem formulation, method design, and bias analysis in Appendix~\ref{appendix_notation}. We build our talent recommendation system based on job postings and resumes of talents, which are organized in a semi-structured way. For example, a resume usually contains textual descriptions of educational background, job experience, project experience, and some keywords to highlight skills, awards, and so on. Thus, we view each job posting and resume as a set of textual content. 

\begin{definition}[Talent Recommendation]
Let $\mathcal{G}$ be the set of job postings and $\mathcal{R}$ be the set of resumes. The historical matching records for jobs are represented by $\mathcal{M}=\{(g_i,r_j,y_{g_i,r_j})\mid g_i\in\mathcal{G},r_j\in\mathcal{R}\}$, where $y_{g_i,r_j}\in\{0,1\}$ and $y_{g_i,r_j}=1$ means $r_j$ matches the requirements of $g_i$. For a new job position $g$ without any matching records, let $\mathcal{C}_g=\{c_1^g,c_2^g,\cdots,c_N^g\}$ be a set of candidate resumes for recommendation, where $c_j^g\in\mathcal{R}$ and $N<\left|\mathcal{R}\right|$. The purpose of talent recommendation is to learn a recommender $F$ bases on a subset (i.e., training set) of $\mathcal{M}$ that can recommend the most suitable resumes from $\mathcal{C}_g$ for $g$.
\end{definition}


In previous attempts for this task, $F$ is a function to measure the matching degree of a job and a resume, making pointwise recommendations based on the predicted result of $g$ and each candidate's resume $c_j^g$, i.e.,
\begin{align}\label{eq1}
    \hat{y}_j^g = F(g, c_j^g), \quad \forall c_j^g\in \mathcal{C}_g
\end{align}
However, pointwise recommendation requires $F$ to repeatedly process $g$ for $N$ times, which is particularly inefficient and resource-intensive, especially when $F$ is an LLM. In addition, pointwise approaches only consider the relationship between $g$ and each resume, ignoring the relationships among resumes within the candidate set. Therefore, we focus on the listwise setting for more efficient and accurate recommendations (\textbf{listwise talent recommendation}), where $F$ takes $g$ and the entire set of candidate resumes $\mathcal{C}_g$ as input and predicts the matching scores for all resumes, i.e., 
\begin{align*}
    \hat{\mathbf{y}}_g = F(g,\mathcal{C}_g) = \left[\hat{y}_1^g, \hat{y}_1^g, \cdots, \hat{y}_N^g \right] 
\end{align*}
We further introduce a ranking function that sorts the candidates in descending order of predicted scores:
\begin{align*}
    \tau_g = \texttt{argsort}(\hat{\mathbf{y}}_g, \text{descending}) 
\end{align*}
The ranked list of candidates is then written as:
\begin{align*}
    \underline{\mathcal{C}}_g  = <c_{\tau_g({1})}^g, c_{\tau_g({2})}^g, \cdots, c_{\tau_g({N})}^g>
\end{align*}
where $\tau_g(l)$ denotes the index of the candidate in $\mathcal{C}_g$ ranked at position $l$.

\section{Methodology}\label{sec4}
In this section, we first introduce the overall procedure for LLM-based list-wise talent recommendation and discuss two methods to extract recommendation results from LLMs. Then, we elaborate on our modified attention mechanism and positional encodings. 

\subsection{Recommendation Process Overview}
\begin{figure}[!htbp]
  \centering
    \includegraphics[width=1\linewidth]{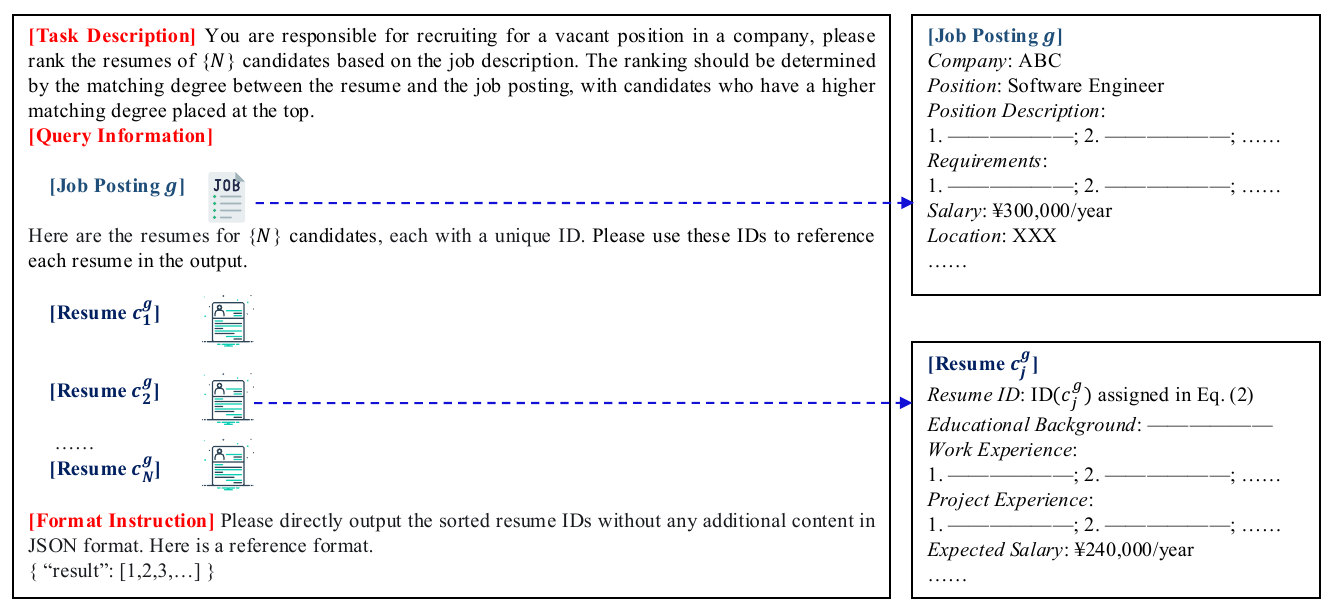}
  \caption{The Prompt Template for Listwise Talent Recommendation.}
  \label{fig1}
\end{figure}

In-context learning (ICL), first proposed with GPT-3~\citep{DBLP:conf/nips/BrownMRSKDNSSAA20}, is a common technique for eliciting the ability of LLMs. ICL typically utilizes a structured prompt in natural language, commonly comprising a task description, a few task samples as demonstrations, and query information. It relies solely on the knowledge and capabilities inherent in LLMs. However, even one task sample in our listwise setting is a very long text (might be longer than 10k), including a job posting and a sequence of resumes. Integrating task samples within the prompt might exceed the context length of LLMs and significantly increase the computational cost. Therefore, we follow the zero-shot ICL method and design the prompt template without demonstrations for listwise talent recommendation. We add format instructions at the end of the prompt to guide the LLM to generate responses that adhere to a specified format. Figure~\ref{fig1} depicts the prompt template, consisting of the task description, the query information including the job posting and the set of candidate resumes, and the format instructions.

Filling the content of job posting $g$ and resumes $\mathcal{C}_g$ into the prompt template, we get the prompt $x_g$, consisting of $n_g$ tokens: $x_1^g,x_2^g,\cdots x_{n_g}^g$,  
\begin{align}\label{eq2}
    x_g = \texttt{BuildPrompt}\left(g,\mathcal{C}_g; \texttt{assign}\left(\mathcal{C}_g\right)\right) = <x_1^g,x_2^g,\cdots x_{n_g}^g> 
\end{align}
where $\texttt{assign}\left(\mathcal{C}_g\right)$ is an operator to assign IDs to resumes in $\mathcal{C}_g$ and $\texttt{BuildPrompt}(\cdot)$ is a prompt generator. A straightforward design of $\texttt{assign}\left(\mathcal{C}_g\right)$ is to sequentially number the resumes in $\mathcal{C}_g$, i.e., $\texttt{assign}(\mathcal{C}_g) =[\text{``1''},\text{``2''},\cdots, \text{``$N$''}]$. Given the well-organized prompt $x_g$, an LLM $F_{\mathrm{LLM}}$ can select the IDs of suitable resumes by generating textual outputs as follows.
\begin{align*}
    o_g = F_{\mathrm{LLM}}(x_g) = <o_1^g, o_2^g, \cdots, o_{|o_g|}^g>
\end{align*}
Most of the LLMs auto-regressively generate the text. They first calculate the output probability of different tokens in the first place and determine the token through a sampling method. This process repeats iteratively until a termination condition is satisfied. In this case, the desired output of a recommender is a list of sorted resume IDs (or a list of matching scores), requiring an additional step to parse the results from textual outputs. Since this generate-then-parse method extracts information directly from the output text that is comprehensible to humans, we refer to it as the explicit recommendation method, formally stated as follows.
\begin{definition}[Explicit Recommendation]\label{def2}
Given the prompt $x_g$ related to the job $g$ and the candidate set $\mathcal{C}_g$, an LLM-based recommender makes explicit recommendations by first generating textual outputs and then parsing the recommendation results.
\end{definition}

The explicit method heavily relies on the instruction-following ability of LLMs and can be widely applied to both open-source LLMs (e.g., Llama~\citep{touvron2023llama}) and the APIs of closed-source LLMs (e.g., ChatGPT~\citep{chatgpt}). Nevertheless, we may fail to parse the results from the text, which inevitably contains various errors, such as duplicate or incorrect IDs of resumes. Fortunately, for open-source LLMs, we can directly access the output probability of every token in the vocabulary, including the ID tokens of candidate resumes when the model generates the first token $o_1^g$. This allows us to make recommendations purely based on these probabilities. Since we do not utilize the generated text, we call this method the implicit recommendation method, as stated below.

\begin{definition}[Implicit Recommendation]\label{def3}
Given the prompt $x_g$ related to the job $g$ and its candidate set $\mathcal{C}_g$, a recommender based on an open-source LLM predicts the matching scores between $g$ and $c_j^g\in\mathcal{C}_g$ based on the output probability of different tokens when it generates the first token $o_1^g$ in the output, i.e., 
\begin{align}\label{eq3}
\hat{y}_j^g = \frac{\mathbb{P}\left( \texttt{\textup{ID}}(c_j^g  )\mid x_g \right)}{\sum\limits_{s=1}^{N} \mathbb{P}\left(\texttt{\textup{ID}}(c_s^g  )\mid x_g \right)},\quad \forall c_j^g \in \mathcal{C}_g
\end{align}
where $\texttt{\textup{ID}}(c_j^g)$ is the assigned ID of $c_j^g$ and $\mathbb{P}\left( \texttt{\textup{ID}}(c_j^g  ) \mid x_g \right)$ is the probability of generating $\texttt{\textup{ID}}(c_j^g)$ as the first token in the output. Then recommendations are made according to the probability scores $\hat{\mathbf{y}}_{g}=\left[{\hat{y}}_1^g,{\hat{y}}_2^g,\cdots,{\hat{y}}_N^g\right]$. For example, we can recommend the top-$k$ ranked resumes. 
\end{definition}

The implicit method only calculates the output probability in the first place, avoiding the time-consuming generation process and the complicated parsing step, which is supposed to be more efficient. Besides, the implicit method doesn’t suffer from the potential errors in the explicit method. Nonetheless, implicit methods need accessibility of the output probability, which is infeasible for closed-source LLMs.

\subsection{Fine-tuning LLMs for Listwise Talent Recommendation}
Although LLMs show promising results on several recommendation tasks~\citep{DBLP:conf/recsys/DaiSZYSXS0X23}, the mismatch between recommendation and the language model task hinders the performance of LLM-based recommenders~\citep{DBLP:conf/recsys/BaoZZWF023,DBLP:journals/tois/ZhangXHZLW25}. Thus, we utilize fine-tuning to bridge the gap. 

For the explicit recommendation methods defined in Definition \ref{def2}, the generation process inevitably introduces unexpected errors, while the subsequent parsing operations are both complex and non-differentiable. As a result, it is challenging to align the recommendation loss function with either the text output or the parsed results. Therefore, we adopt the implicit recommendation methods defined in Definitions \ref{def3} for fine-tuning.

To fine-tune open-source LLMs with our task, we arrange the available matching records $\mathcal{M}$ to mimic the listwise setting. For each job $g_i\in\mathcal{G}$ that has matched resumes in $\mathcal{M}$, the set of resumes $\mathcal{R}$ can be split into a positive set $\mathcal{R}_{g_i}^+$ and a negative set $\mathcal{R}_{g_i}^-$ according to the matching records. Since the number of resumes directly impacts the prompt length, and the resource requirements (e.g., GPU vRAM) during training are substantially higher than those during inference, we set the number of resumes in the training prompt, $N_{\mathrm{train}}$, to be smaller than $N$, the number of resumes used during inference. We leverage each positive record to form the candidate set for each training sample. Specifically, for each record $\left(g_i,r_k,y_{g_i,r_k}=1\right)\in\mathcal{M}$, the candidate set $\mathcal{C}_{g_i,r_k}$ contains the positive resume $r_k$ and $(N_{\mathrm{train}}-1)$ resumes randomly sampled from $\mathcal{R}_{g_i}^-$. 

As $N_{\mathrm{train}}<N$, IDs of some resumes during inference may not have appeared during training. The assigned ID of each resume is crucial when calculating the matching score based on Eq. (\ref{eq3}). Therefore, this mismatch issue ($N_{\mathrm{train}}<N$) poses challenges for LLMs to predict the matching scores of resumes whose IDs do not appear during training. To solve this problem, we introduce the following sampling procedure when assigning IDs of resumes in Eq. (\ref{eq2}). 
\begin{align}\label{eq4}
    \texttt{assign}\left(\mathcal{C}_{g_i,r_k}\right) = \texttt{sample}\left(\mathcal{I}, N_{\mathrm{train}}\right)
\end{align}
where $\texttt{sample}(\mathcal{I}, N_{\mathrm{train}})$ is an operator that uniformly samples $N_{\mathrm{train}}$ unique IDs from a large set of IDs $\mathcal{I}$. This assignment method ensures that the LLM can learn information of a sufficiently diverse set of IDs during training, so as to be able to handle candidate sets longer than $N_{\mathrm{train}}$ during inference. Additionally, because the candidate resume IDs in each training sample are randomly chosen and differ across samples, this variability helps prevent the LLM from acquiring a preference for specific IDs (token bias), steering its attention toward the underlying content. 

For the training sample built on record $\left(g_i,r_k,y_{g_i,r_k}=1\right)\in\mathcal{M}$, we obtain the predicted matching score ${\hat{y}}_j^{g_i}$ of each resume $c_j \in\mathcal{C}_{g_i,r_k}$ based on the output probability of the first token, as shown in Eq. (\ref{eq3}). Then we define the loss function for this training sample as follows.
$$
\mathcal{L}_{g_i,r_k} = \sum_{c_j \in\mathcal{C}_{g_i,r_k}} -y_{g_i,c_j} \log \hat{y}_j^{g_i}
$$
where $y_{g_i,c_j}\in\{0,1\}$ is the true matching result of $c_j$ with job position $g_i$.

\subsection{Block Attention}
In talent recommendations, LLMs should primarily focus on the relationship between the job posting and each candidate resume. Simultaneously, LLMs should also consider the relationships among the candidates. As shown in Figure~\ref{fig1}, the prompt $x$ for a job posting $g$ (omitting subscript or superscript where no ambiguity arises) begins with the task description and the job posting, followed by each resume in the candidate set. Based on this structure, we divide $x$ into several blocks, as shown below.
\begin{align}\label{eq5}
    x = [\underset{\text{task description}}{\underbrace{x_1,\cdots,x_{b_0}}},\ \underset{\text{job posting }g}{\underbrace{x_{b_0+1},\cdots,x_{b_1}}},\ 
    \cdots,\ \underset{\text{resume }c_j}{\underbrace{x_{b_j+1},\cdots,x_{b_{j+1}}}},\ \cdots]
\end{align}
The task description is viewed as the zeroth block containing $b_0$ tokens. The job posting is the first block with $b_1-b_0$ tokens, while the $j$-th resume $c_j$ is the $j+1$-th block with $b_{j+1}-b_j$ tokens. 

\begin{figure}[!htbp]
    \centering
    \includegraphics[width=0.5\linewidth]{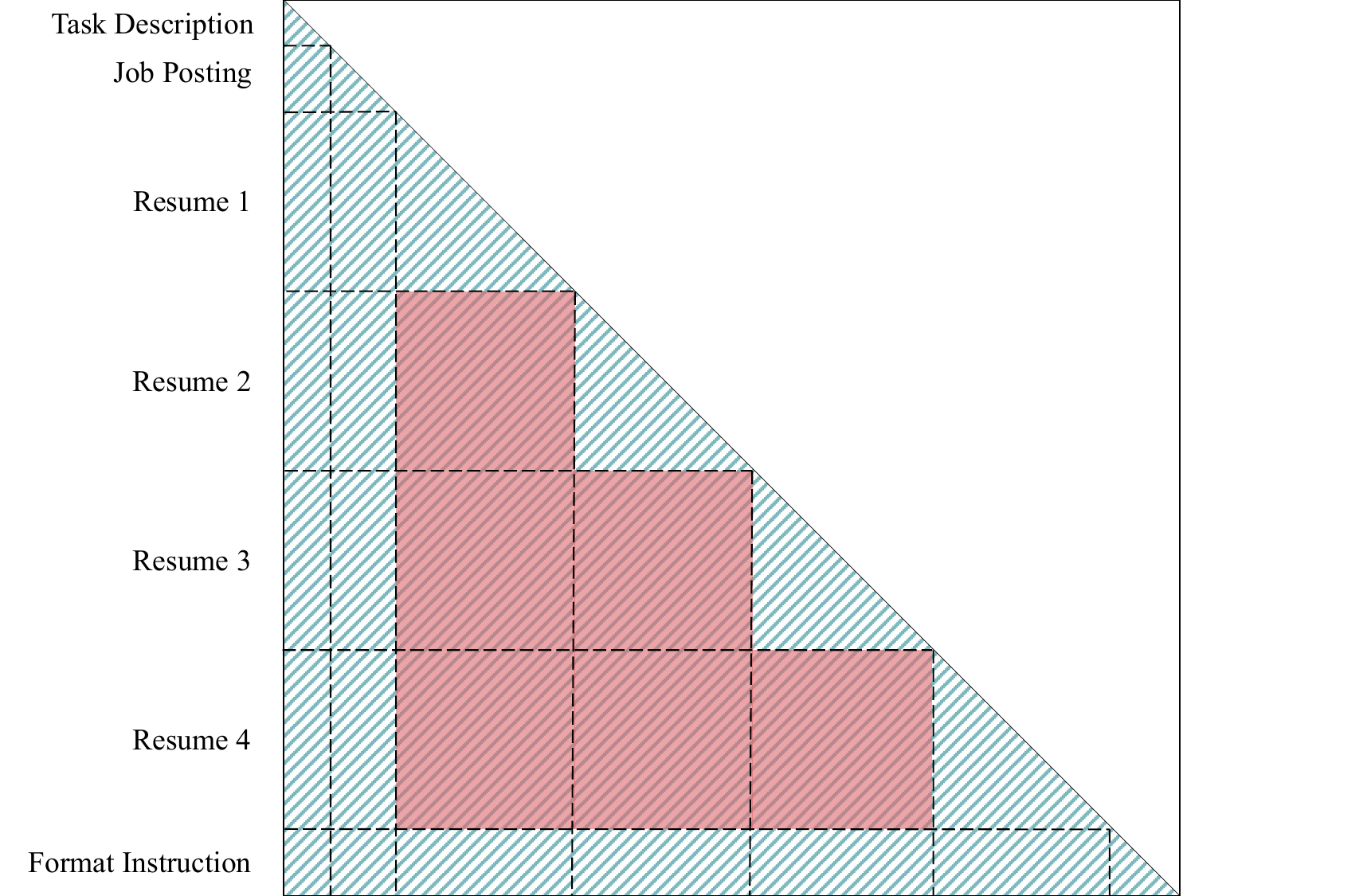}
    \caption{Causal Attention Map.}
    \label{fig2}
\end{figure}

 Figure~\ref{fig2} displays the attention map of a causal decoding LLM\footnote{Most modern large language models adopt the causal decoder architecture of the Transformer, including representative models such as the GPT series~\citep{DBLP:conf/nips/BrownMRSKDNSSAA20,chatgpt}, LLaMA series~\citep{touvron2023llama,grattafiori2024llama}, and GLM series~\citep{glm2024chatglm,zengglm}}, taking the processing of one job posting and four resumes as an example. Rows in the figure represent the current position (Query) we are computing attention for, and  columns represent the positions we are allowed to look at (Key). The red part highlights the attention calculations between resume tokens.

Listwise recommendation methods have the advantage of modeling the interdependencies among candidate resumes and avoiding redundant processing of the same job posting. However, the default causal attention mechanisms treat all tokens equally. This indiscriminate full attention leads to two issues: (i) substantial computational overhead due to quadratic interactions among all resume tokens, and (ii) dilution of meaningful job–resume and resume–resume relationships. In particular, the model may overemphasize local details within individual resumes, while overlooking the higher-level relational structure across resumes.

\begin{figure}[!htbp]
  \centering
  \subfigure[Block Attention.]{\label{fig3a}
  \includegraphics[width=0.44\linewidth]{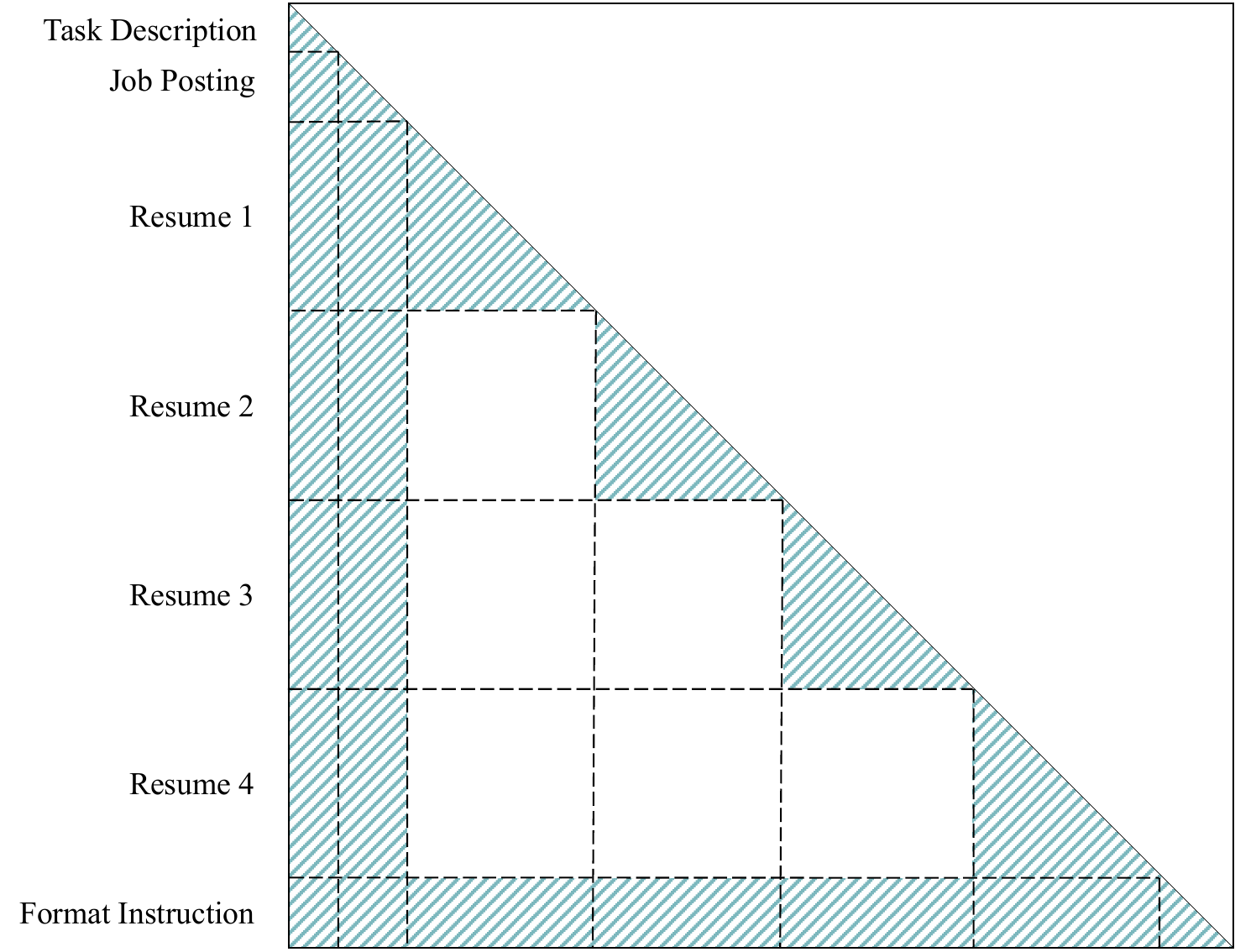}
  }
  \subfigure[Block Attention with Special Tokens.]{\label{fig3b}
  \includegraphics[width=0.44\linewidth]{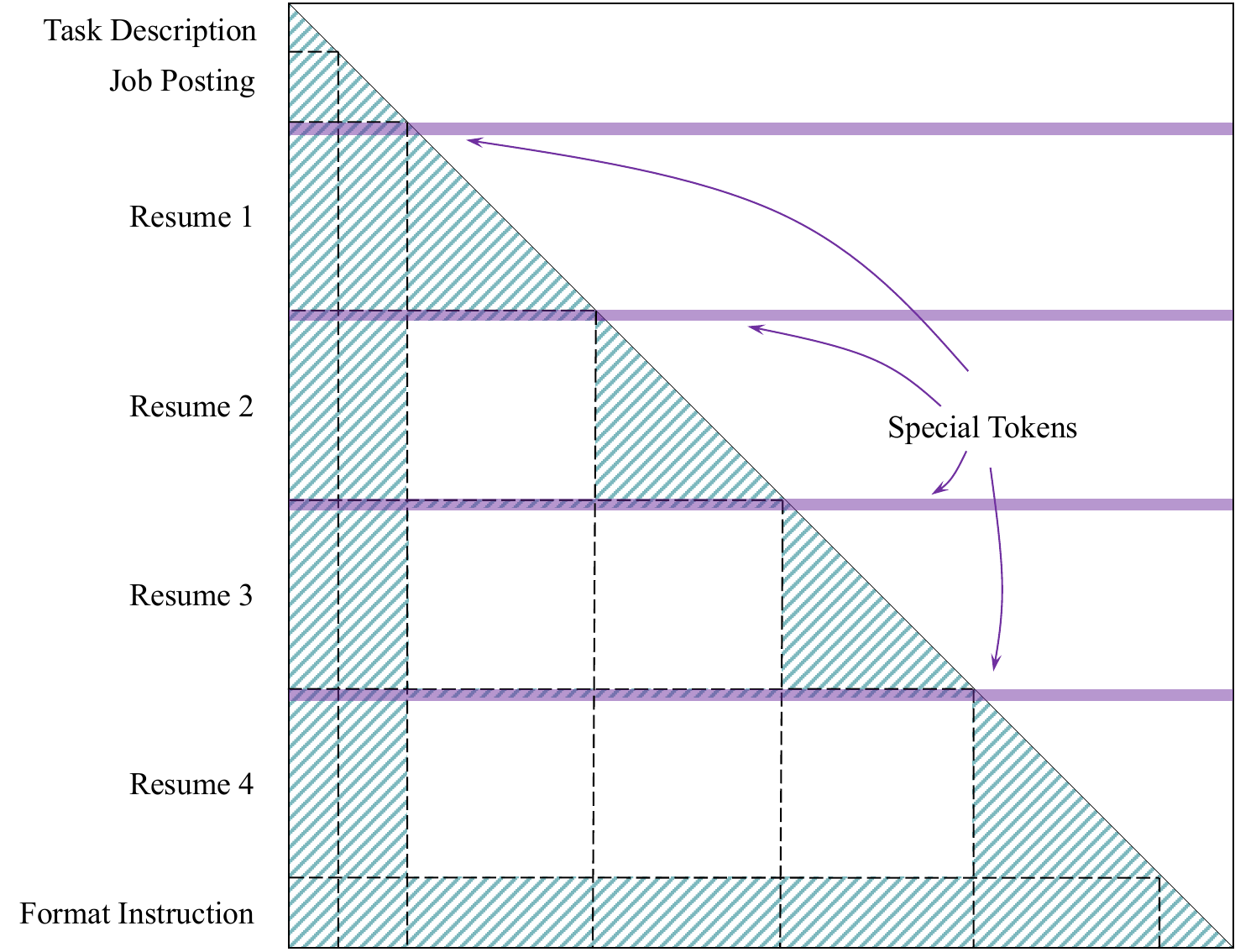}
 }
  \caption{Attention Map of Block Attention and Its Variant.}
  \label{fig:simulation}
\end{figure}

We propose a modified block attention mechanism that can reduce the cumbersome computations among resumes while ensuring that LLMs do not overlook the relationships among them. We first remove the attention among tokens in different resumes, which is almost equivalent to the pointwise setting, as indicated by the blank cell in Figure~\ref{fig3a}. For each resume $c_j$, we prepend a learnable token $x^{c_j}$. The embedding of this token is initialized using the average token embedding of the corresponding resume and is further fine-tuned during training. In addition, we designate the first 
$S$ tokens of each resume as special tokens, which include the inserted learnable token. These special tokens can perform attention calculations with every token in the prompt, as shown in the attention map in Figure~\ref{fig3b}. The remaining normal tokens are restricted to attending only to tokens within the same resume. We expect these special tokens to represent the comprehensive information of their respective resumes, and through the attention computation between special tokens, LLMs can compare different resumes and produce better ranking results. 

\begin{equation}\label{eq6}
\begin{split}
    x' & = [\underset{\text{task description}}{\underbrace{x_1,\cdots,x_{b_0}}},\ \underset{\text{job posting }g}{\underbrace{x_{b_0+1},\cdots,x_{b_1}}},\ 
    \cdots,\ \underset{\text{learnable tokens for }c_j}{\underbrace{x^{c_j}}},\  \underset{\text{resume }c_j}{\underbrace{ x_{b_j+1},\cdots,x_{b_{j+1}}}},\ \cdots] \\
    & \overset{\text{remap indices}}{=} [\underset{\text{task description}}{\underbrace{x_1,\cdots,x_{b_0}}},\ \underset{\text{job posting }g}{\underbrace{x_{b_0+1},\cdots,x_{b_1}}},\ 
    \cdots,\ \underset{\text{$S$ special tokens of resume }c_j }{\underbrace{x_{b_j+1},\cdots,x_{b_j+S}}}, \underset{\text{Normal tokens of resume }c_j }{\underbrace{x_{b_j+S+1},\cdots,x_{b_{j+1}}}} \ \cdots]
\end{split}
\end{equation}

\subsection{Local Positional Encoding}
Nearly all state-of-the-art LLMs (e.g., GPT, BERT, Llama) are built on the Transformer architecture \citep{DBLP:conf/nips/VaswaniSPUJGKP17}, and self-attention is the Transformer’s defining component, enabling LLMs to model contextual and long-range linguistic relationships. However, it cannot recognize the order of the input sequence, thus requiring explicit positional information to be injected, typically in the form of positional encodings. Common positional encoding methods such as RoPE \citep{DBLP:journals/ijon/SuALPBL24} often exhibit {\bf long-term decay}, which means the inner product (the core operation in self-attention) decays when the relative distance between tokens increases. While this property aligns to some extent with the intuitive notion of "paying more attention to nearby tokens", excessive decay can severely impair the model's ability to process long sequences and capture long-range dependencies. The long-term decay inherently introduces a form of position bias, which refers to the model’s tendency to assign disproportionate attention to content appearing earlier or closer in the input sequence, regardless of its actual semantic relevance. In structured prompts or multi-part inputs, as shown in Eq. (\ref{eq5}), such bias may cause the model to overemphasize the sections positioned near the query while underutilizing distant but relevant ones \citep{liu2024lost}. In our context, this property weakens the interaction between the job posting and the distant resumes, thereby amplifying this position bias and constituting a severe constraint on modeling capabilities.

For the listwise talent recommendations task, the tokens in the job posting block are important for each resume. However, the long-term decay property hinders the attention calculation between the last few resumes and the job postings due to the long distance. To solve this problem, we devise a local positional encoding method to mitigate the decay effect.

In LLMs, the text input $x'$ (the prompt with special tokens) is first tokenized, and each token is represented as an embedding vector through an embedding layer, i.e.,
$$
\mathbf{x'} = \texttt{emb}(x') = [\mathbf{x}_1, \cdots , \mathbf{x}_{b_0},\cdots, \mathbf{x}_{b_0+1},\cdots,\mathbf{x}_{b_1}, \cdots , \mathbf{x}_{b_j+1}, \cdots, \mathbf{x}_{b_{j+1}}, \cdots]
$$
where $\texttt{emb}(\cdot)$ returns the embedding of each token and $\mathbf{x}_i$ is a $d$-dimensional vector. To explicitly inject position information, an embedding for each position $i$ is usually obtained through a positional encoding function, i.e., $\boldsymbol{p}_i = \texttt{PE}(i,d)$, where $d$ controls the dimension of position embeddings and should be aligned with that of $\mathbf{x}_i$. Then each token $\mathbf{x}_i$ is mapped into queries $\boldsymbol{q}_i$, keys $\boldsymbol{k}_i$, and values $\boldsymbol{v}_i$ as follows. 
\begin{align}\label{eq7}
    \boldsymbol{q}_i  = f_q(\mathbf{x}_i, \boldsymbol{p}_i), \quad \boldsymbol{k}_i  = f_k(\mathbf{x}_i, \boldsymbol{p}_i),\quad \boldsymbol{v}_i  = f_v(\mathbf{x}_i, \boldsymbol{p}_i)
\end{align}
where $\boldsymbol{q}_i$, $\boldsymbol{k}_i$, and $\boldsymbol{v}_i$ incorporate the $i$-th position through functions $f_q$, $f_k$, and $f_v$, respectively. Different position encoding methods utilize distinct $\texttt{PE}(\cdot,\cdot)$ to yield different $\boldsymbol{p}_i$. For example, RoPE generates a sparse rotary matrix and converts it to two dense vectors for element-wise multiplication. The queries and keys are utilized to calculate attention weights, which are used to get the weighted sum over the values, the output of the self-attention layer.

Considering the structure of the prompt, we split $\mathbf{x'}$ into several blocks as shown in Eq. (\ref{eq6}). From the global view of the whole sequence, the position information $\boldsymbol{p}_i$ of $\mathbf{x}_i$ can be injected as shown in Eq. (\ref{eq7}), enabling the model to learn the relationship between any token and $\mathbf{x}_i$. To capture the connection between $c_j$ and the job posting $g$ and mitigate the effect of the long-term decay problem, we introduce a local view that only contains $c_j$ and $g$, equivalent to placing $c_j$ directly after $g$ in the prompt. Then for each token $\mathbf{x}_i$ that belongs to the resume $c_j$, we calculate local position embeddings $\boldsymbol{p}_i'$ by dropping all tokens from the second to the $j$-th block in the original prompt: 
\begin{align}\label{eq8}
    \boldsymbol{p}_i' = \texttt{PE}(i-b_j+b_1, d)
\end{align}

\begin{figure}[!htbp]
  \centering
    \includegraphics[width=0.9\linewidth]{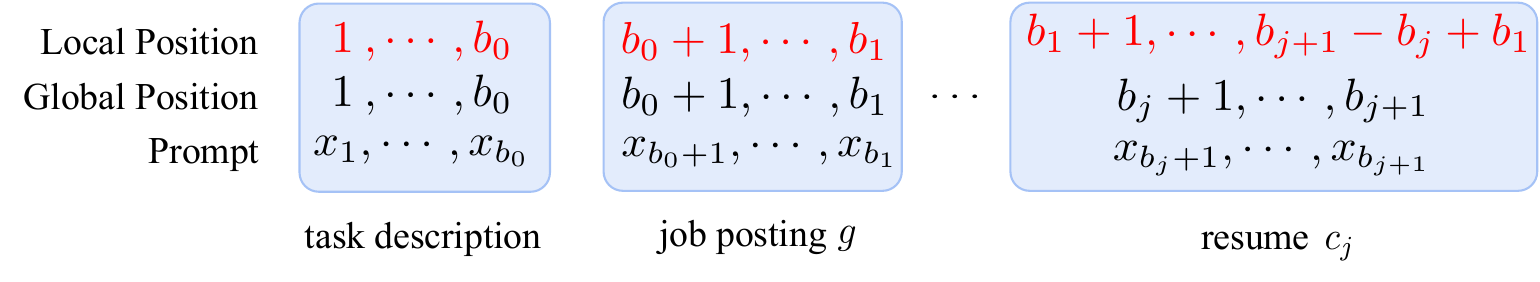}
  \caption{Illustration of Global and Local Position Information.}
  \label{fig3}
\end{figure}

Figure \ref{fig3} shows the global and local positions of ${x'}$, where position of ${x}_{b_j+1}$, i.e., the first token of resume $c_j$, becomes $b_1+1$.

We utilize the local position information by replacing $\boldsymbol{p}_i$ in Eq. (\ref{eq7}) with $\boldsymbol{p}_i'$. This replacement can enhance the attention calculation between tokens in the job posting and each resume since it reduces the position distance, which will mitigate the negative effect of the long-term decay property. Besides, the special tokens corresponding to different resumes share identical positional information (e.g., $x_{b_j+1}$ and $x_{b_{j+1}+1}$ in Eq. (\ref{eq6})), and they can attend to each other, which means that the computation among these special tokens is unordered. This design is consistent with the fundamental property of listwise recommendation, where the candidate list is conceptually a permutation-invariant set, i.e., the model’s ranking results should remain unchanged regardless of the input order of resumes. We provide theoretical analysis of the local position encoding in Appendix~\ref{appendix_a}.

Figure~\ref{fig5} shows the architecture of our proposed LLM-based recommendation model, where the blue font highlights our special designs. We utilize a common strategy, e.g., LoRA~\citep{DBLP:conf/iclr/HuSWALWWC22}, to fine-tune the LLM, which inserts additional learnable parameters at each transformer layer while keeping other parameters fixed. 

\begin{figure}[!htbp]
    \centering
    \includegraphics[width=0.75\linewidth]{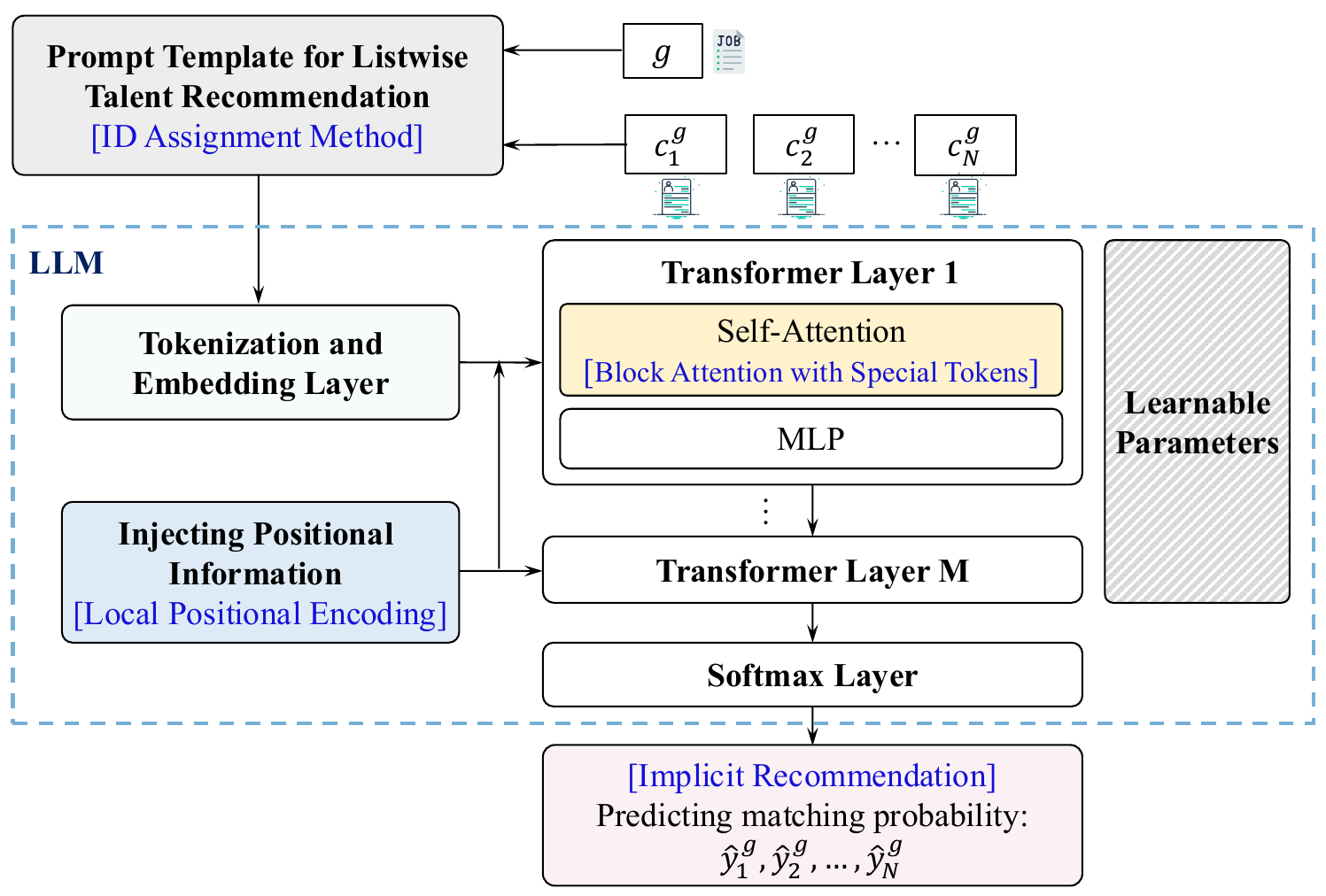}
    \caption{Architecture of Our Proposed LLM-based Recommendation Model.}
    \label{fig5}
\end{figure}

\section{Debiasing Method}\label{sec5}
LLM-based recommenders show volatile performance under the permutation of the candidate list ~\citep{hou2024large}. In this section, we first investigate the bias patterns and propose methods to detect position and token bias. Then, we design two methods to mitigate the bias issue: a pre-ranker to utilize the bias pattern and a probability-based debiasing method.

\subsection{Bias Patterns and Bias Estimation}\label{sec5.1}
To begin with, we investigate the bias patterns that may arise in listwise recommendation settings. Using a real-world talent-job matching dataset, HRT (see Section~\ref{sec6.1} for more details), we set the length of the candidate list to $N=15$. Each candidate resume in the prompt is assigned a unique ID ranging from ``A'' to ``O''. We randomly select a subset $\mathcal{M'}$ from all matching records $\mathcal{M}$ to construct a test dataset $\mathcal{T}$. For each selected positive job–resume pair $(g, r)$, we build a candidate list $\mathcal{C}_g$ for the job posting $g$ by including the matched resume $r$ and 14 resumes randomly sampled from those not associated with $g$. We denote the ordered version of this candidate list as $\mathbf{C}_g = \langle c_1^g, c_2^g, \cdots, c_N^g \rangle$, where candidates are alphabetically sorted by their names by default. The overall test set can thus be expressed as $\mathcal{T} = \{\mathbf{C}_g \mid (g, r, 1) \in \mathcal{M}'\}$. To measure the impact of the matched resume's position on the recommendation performance, we manipulate its position in the candidate list.  Specifically, let the matched resume $r$ initially occupy position $Q$. We then vary $Q$ across all other positions, while keeping the relative order of the remaining 14 resumes unchanged.

\begin{figure}[!htbp]
    \centering
    \includegraphics[width=0.5\linewidth]{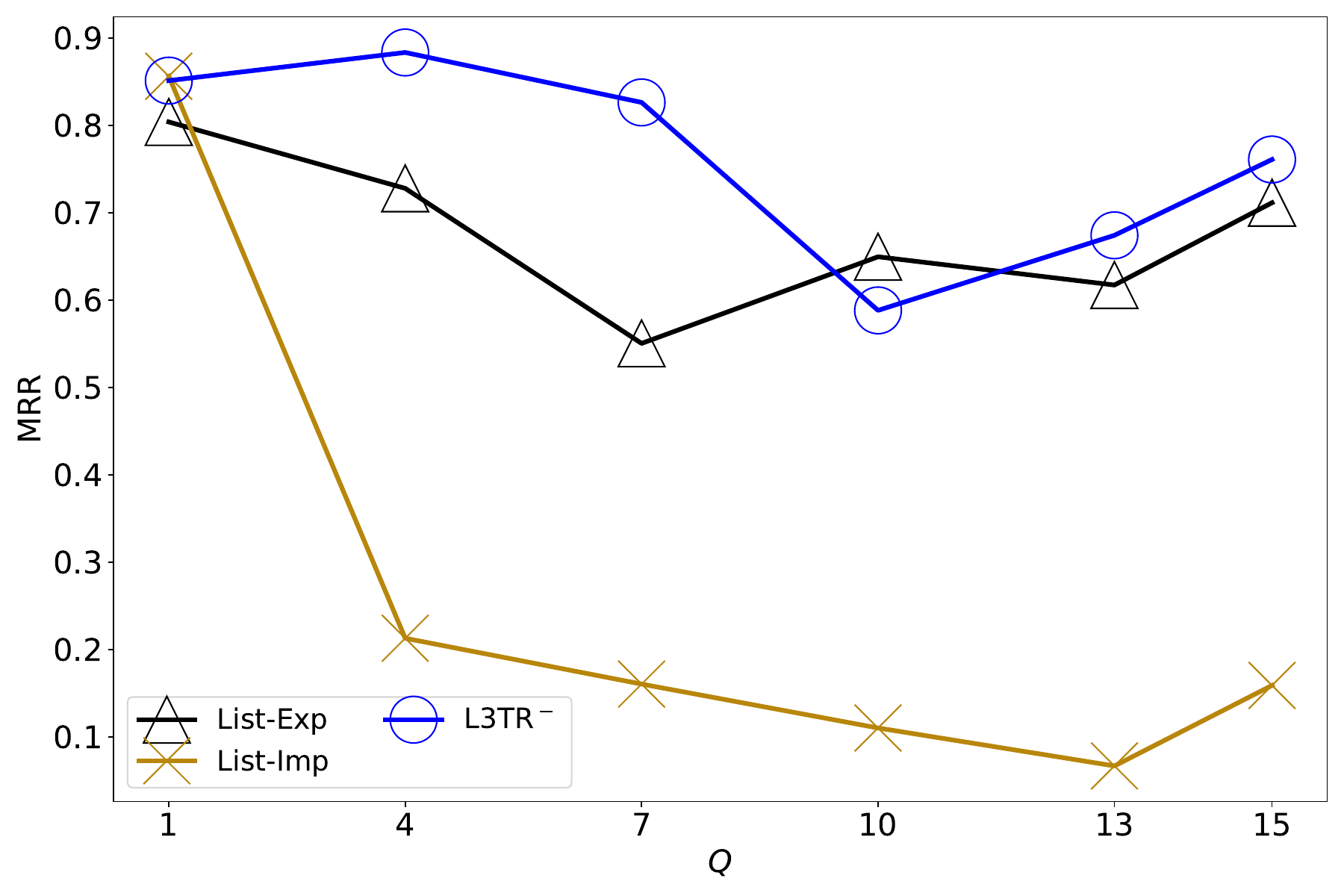}
    \caption{Performance with Different $Q$.}
    \label{fig6}
\end{figure}

Figure \ref{fig6} shows the Mean Reciprocal Ranking (MRR) for various models with different $Q$. List-Imp leverages an open-source LLM (i.e., ChatGLM3-6B~\citep{zengglm}) and adheres to the implicit method in Definition~\ref{def3}. List-Exp performs explicit listwise recommendations with a large commercial closed-source LLM (i.e., GLM-4~\citep{glm2024chatglm}). L3TR$^-$ utilizes listwise data and the LoRA method to fine-tune ChatGLM3-6B without any special designs proposed in this paper. We can observe that List-Imp and even List-Exp with over 100B parameters are quite sensitive to $Q$. Although the fine-tuned LLM can gain better overall scores, it still exhibits unstable performances. This observation is largely consistent with the results reported in other literature ~\citep{hou2024large,liu2024lost}.

These bias patterns might be attributed to the {position bias} and the {token bias}~\citep{zheng2023large}. \textbf{Position bias} refers to the tendency of LLMs to favor content that is placed at specific positions, such as the first or second position in the candidate list. \textbf{Token bias} refers to LLMs' prior tendency to assign more probability mass to certain ID tokens, such as ``B'' or ``D''. \citet{zheng2023large} propose some methods to evaluate the position and token bias when LLMs are required to answer multiple-choice questions (MCQs). For example, \citet{zheng2023large} remove ID tokens in MCQs and prompt LLMs to output the option contents. This way can eliminate the impact of token bias, leaving only position bias. However, in listwise talent recommendations, it's impractical to expect LLMs to output the entire selected resume due to its excessive length. Therefore, we propose more reasonable and practical methods to estimate the position and token bias in our scenario.
 
Let $\mathcal{D} = \{d_1,d_2,\dots,d_N\} = \{\text{``A''},\text{``B''},\dots,\text{``O''}\}$ denote the set of ID tokens assigned to the ordered candidate list $\mathbf{C}_g$ in the test set $\mathcal{T}$.  Let $\lambda \in \Lambda(\mathbf{C}_g)$ represent a permutation of the candidate list $\mathbf{C}_g$, and let $\pi \in \Pi(\mathcal{D})$ denote a permutation of the ID token set $\mathcal{D}$. We focus on measuring {token bias}, defined as the inherent preference of the LLM for each ID token $d_i \in \mathcal{D}$. To make the preference comparable across different IDs and mitigate the influence of the surrounding IDs (e.g., contrast effects), we introduce a distinct \emph{reference ID set} $\mathcal{B}$ with size $|\mathcal{B}|=|\mathcal{D}|-1$.  For each $\mathbf{C}_g$, exactly one resume is assigned the target ID $d_i$, while the remaining $N-1$ resumes are assigned IDs from $\mathcal{B}$. For example, when $N=15$ and the target ID is $d_i=\text{``A''}$, we can choose $\mathcal{B}=\{\text{``a''},\text{``b''},\dots,\text{``n''}\}$. A possible ID assignment for 
$\mathbf{C}_g$ is 
$$\langle \text{``A''}: c_1^g, \ \text{``a''}: c_2^g,\  \text{``b''}: c_3^g,\  \cdots,\  \text{``n''}: c_{15}^g \rangle$$
resulting in the ID set $\mathcal{B}_A = \{\text{``A''},\text{``a''},\text{``b''},\dots,\text{``n''}\}$. 
The reason for introducing $\mathcal{B}$ is to ensure a controlled comparison across different ID tokens. When comparing tokens $d_i \in \mathcal{D}$, it is important to keep ID tokens of all other positions identical. Otherwise, changing the target token may also change IDs at other positions. For instance, suppose we compare ``A'' and ``C'' at the first position, while filling the remaining positions with tokens from $\mathcal{D}$ in alphabetical order. The resulting ID sequences $\{\text{``A''},\text{``B''},\text{``C''},\cdots, \text{``O''}\}$ and $\{\text{``C''},\text{``B''},\text{``A''},\cdots, \text{``O''}\}$ differ not only at the first position (``A'' vs. ``C'') but also at the third position (``C'' vs. ``A''). Such unintended differences introduce additional variation that may confound the comparison. By introducing the reference set $\mathcal{B}$, all non-target positions can be fixed with the same reference IDs in a consistent order, ensuring that the comparison isolates the effect of the target token.

Under this setup, we exhaustively vary both the order of the candidates ($\lambda \in \Lambda(\mathbf{C}_g)$) and the order of ID assignments ($\pi \in \Pi(\mathcal{B}_{d_i})$) to enumerate all possible ID–resume combinations.   Now the preference of LLMs for the target token $d_i \in \mathcal{D}$ can be expressed as 
\begin{align}\label{eq9}
\mathbb{P}(d_i\mid \mathcal{B}_{d_i}) = \frac{1}{|\mathcal{T}|}\sum_{\mathbf{C}_g\in\mathcal{T}}\frac{1}{|\Pi(\mathcal{B}_{d_i})|\cdot |\Lambda(\mathbf{C}_g)|}\sum_{\pi\in\Pi(\mathcal{B}_{d_i})}\sum_{\lambda\in\Lambda(\mathbf{C}_g)}\mathbb{P}\left( d_i \mid g, \pi, \lambda, \mathcal{B}_{d_i}, \mathbf{C}_g\right)
\end{align}
where $\mathbb{P}(d_i\mid \cdots )$ is the probability that the LLM selects token $d_i$. 

It's infeasible to enumerate each $\pi$ and $\lambda$ due to the time and computation constraints ($\times N!$ costs). To make the procedure tractable, we restrict our analysis to a special set of permutations. First, for $\pi$ we consider
\begin{align*}
    \pi^{*}_j(d_i) = \{\text{``a''}, \text{``b''}, \cdots, \underbrace{d_i}_{\text{Position $j$}}, \cdots ,\text{``n''}\}
\end{align*}
where the target ID $d_i$ is placed in the $j$-th position, and the remaining $N-1$ IDs 
retain their original relative order. Similarly, we also consider special permutations of the candidate resumes $\lambda_j^{*}$, where the matched resume is placed in the $j$-th position and other resumes remain their original order in $\mathbf{C}_g$. That is to say, the target ID $d_i$ is always assigned to the matched resume, and we only vary its position in the list. In this way, LLM's preference for the ID is equivalent to its preference for the matched resume. Thus, we can use the LLM's recommendation accuracy performance to reflect its preference on the target ID,  making it applicable to both open- and closed-source LLMs. Formally, the estimated preference for $d_i$ at position $j$ is given by
\begin{align}\label{eq10}
    \frac{1}{|\mathcal{T}|}\sum_{\mathbf{C}_g\in\mathcal{T}} \mathbb{P}\left( d_i \mid g, \pi^{*}_j(d_i), \lambda_j^{*}, \mathcal{B}_{d_i}, \mathbf{C}_g\right)
\end{align}
Summing over the position yields an approximate overall preference for $d_i$.
\begin{align}
    \tilde{\mathbb{P}}(d_i\mid \mathcal{B}_{d_i}) = \frac{1}{N}\sum_{j=1}^{N}\frac{1}{|\mathcal{T}|}\sum_{\mathbf{C}_g\in\mathcal{T}} \mathbb{P}\left( d_i \mid g, \pi^{*}_j(d_i), \lambda_j^{*}, \mathcal{B}_{d_i}, \mathbf{C}_g\right)
\end{align}
Similarly, we can define the preference of LLMs for position $j$ as follows.
\begin{align}\label{eq12}
    \mathbb{P}(\text{position } j \mid \mathcal{D}) = \frac{1}{|\mathcal{T}|}\sum_{\mathbf{C}_g\in\mathcal{T}}\frac{1}{|\Pi(\mathcal{D})|\cdot |\Lambda(\mathbf{C}_g)|}\sum_{\pi\in\Pi(\mathcal{D})}\sum_{\lambda\in\Lambda(\mathbf{C}_g)}\mathbb{P}\left( d_{\pi(j)} \mid g, \pi, \lambda, \mathcal{D}, \mathbf{C}_g\right)
\end{align}
where $\pi(j)$ is the index of $j$-th ID in $\pi$ within the original ID token set $\mathcal{D}$. We also set the permutation of the candidate resumes as $\lambda_j^*$. The right-hand side of Eq. (\ref{eq12}) reduces to 
\begin{align*}
    \frac{1}{|\mathcal{T}|}\sum_{\mathbf{C}_g\in\mathcal{T}}\frac{1}{|\Pi|(\mathcal{D})|}\sum_{\pi\in\Pi(\mathcal{D})}\mathbb{P}\left( d_{\pi(j)} \mid g, \pi, \lambda_j^*, \mathcal{D}, \mathbf{C}_g \right)
\end{align*}
Enumerating all ID permutations $\pi$ for each job posting $g$ is still computationally expensive. Therefore, for each job $g$, we randomly sample a single permutation 
$\pi^g \sim \Pi(\mathcal{D})$ so that the distribution of the ID token before each position $j$ is approximately consistent at the level of the entire dataset, which can eliminate the impact of the token bias.
\begin{align}\label{eq13}
    \tilde{\mathbb{P}}(\text{position } j\mid \mathcal{D}) = \frac{1}{|\mathcal{T}|}\sum_{\mathbf{C}_g\in\mathcal{T}} \mathbb{P}\left( d_{\pi^g(j)} \mid g, \pi^g, \lambda_j^*, \mathcal{D}, \mathbf{C}_g\right)
\end{align}
Using these methods, we evaluate the debiasing capability of the proposed model in Section~\ref{sec6.4}.

\subsection{Pre-ranker}\label{sec5.2}
Many LLM-based recommenders tend to achieve higher scores when the matched resume appears at a smaller position index $Q$, i.e., when it is placed near the beginning of the candidate list. Motivated by this observation, we design a pre-ranker to exploit this property. Specifically, we first compute the average word embeddings of the job posting and each resume in the candidate list.
Then, we rank the resumes according to their cosine similarity scores with the job posting. This pre-ranking step is applied before the formal LLM-based recommendation so that potentially matched resumes are placed closer to the front of the prompt, which can leverage the LLM’s preference for the content appearing earlier in the candidate list.

\subsection{Probability-based Debiasing Method}
Token bias and position bias are inherently entangled. The LLM may prefer different ID tokens at various positions. Therefore, we aim to mitigate these two sources of bias simultaneously. The true preference of the LLM over resume $c_i$ can be defined as follows.
\begin{align}
    \mathbb{P}(c_i\mid g, \mathcal{D}, \mathbf{C}_g) = \frac{1}{|\Pi(\mathcal{D})|\cdot|\Lambda(\mathbf{C}_g)|} \sum_{\pi\in\Pi(\mathcal{D})}\sum_{\lambda\in\Lambda(\mathbf{C}_g)} \mathbb{P}(d_{\pi(f_{\lambda}(c_i))}\mid g,\pi,\lambda,\mathcal{D},\mathbf{C}_g)
\end{align}
where $f_{\lambda}(c_i)$ returns the index of $c_i$ under the permutation $\lambda$ of the candidate list. In practice, it's prohibitively expensive to compute all permutations. When calling the LLM to make recommendations, we can instead observe the joint distribution of $c_i$ and an ID token $d_j$ as follows.
\begin{align}\label{eq15}
    \mathbb{P}_{\mathrm{obs}}(c_i,d_j\mid g,\pi,\lambda,\mathcal{D}, \mathbf{C}_g) = \left\{\begin{array}{ll}
			\mathbb{P}_{\mathrm{obs}}\left( d_j\mid g,\pi,\lambda,\mathcal{D},\mathbf{C}_g \right), & j = \pi(f_{\lambda}(c_i)) \\
			0,& \text{otherwise} 
		\end{array}\right.
\end{align}
where $\mathbb{P}_{\mathrm{obs}}\left( d_j\mid g,\pi,\lambda,\mathcal{D},\mathbf{C}_g \right)$ is the output of implicit recommendation in Eq. (\ref{eq3}). We assume that the observed joint distribution of $c_i$ and $d_j$ can be decomposed as a prior distribution over $d_j$ (capturing prior bias) and the true preference distribution over $c_i$. This assumption, referred to as the \textbf{Conditional Independence Assumption}. It's equivalent to saying that the LLM holds independent beliefs about $d_j$ and $c_i$. 
\begin{align}\label{eq16}
    \mathbb{P}_{\mathrm{obs}}\left(c_i,d_j\mid g,\pi,\lambda,\mathcal{D},\mathbf{C}_g\right) = \mathbb{P}_{\mathrm{prior}}\left(d_{\pi(f_{\lambda}\left( c_i \right))}\mid g,\pi,\lambda,\mathcal{D}, \mathbf{C}_g\right)\mathbb{P}\left(c_i\mid g,\pi,\lambda,\mathcal{D}, \mathbf{C}_g\right)
\end{align}
This assumption is reasonable because ID tokens are synthetic identifiers that do not carry semantic information about the resumes. They are only used to mark the positions of candidates within the prompt, and the LLM has no inherent reason to associate a particular ID token with particular resume content. Under this assumption, the true preference for $c_i$ should be invariant to permutations of both resumes and ID tokens, i.e., $\mathbb{P}(c_i \mid g, \pi, \lambda, \mathcal{D},\mathbf{C}_g) = \mathbb{P}(c_i \mid g, \mathcal{D}, \mathbf{C}_g)$. In practice, however, LLM-based recommendation operates on a fixed input order. We consider a canonical ordering where $\tilde{\lambda}$ follows the alphabetical order and $\tilde{\pi} = \{\text{``A''}, \text{``B''}, \cdots\}$, such that $i = \tilde{\pi}(f_{\tilde{\lambda}}(c_i))$. Under this fixed ordering, the prior bias towards $d_i$ is primarily determined by the position of the matched resume. Therefore, $\mathbb{P}_{\mathrm{prior}}$ no longer depends on the specific permutation $\pi$ or mapping $\lambda$, but only on $g$ and $\mathcal{D}$. As a result, Eq. (\ref{eq16}) to be simplified as follows:
\begin{align}
    \mathbb{P}_{\mathrm{obs}}\left(d_i\mid g,\tilde{\pi},\tilde{\lambda},\mathcal{D}, \mathbf{C}_g\right) = \mathbb{P}_{\mathrm{prior}}\left(d_i\mid g, \mathcal{D}, \mathbf{C}_g\right)\mathbb{P}\left(c_i\mid g, \mathcal{D}, \mathbf{C}_g \right)
\end{align}
We can utilize the globally observed bias as an approximation of the prior bias of each $g$
\begin{align}\label{eq18}
    \hat{\mathbb{P}}\left(d_i\mid \mathcal{D}\right) =  \frac{1}{|\mathcal{T}|} \sum_{\mathbf{C}_g\in\mathcal{T}}\mathbb{P}_{\mathrm{obs}}\left(d_i\mid g, \tilde{\pi},\tilde{\lambda}, \mathcal{D}, \mathbf{C}_g\right)
\end{align}
We estimate $ \hat{\mathbb{P}}\left(d_i\mid \mathcal{D}\right) $ on a small development set and then use it for debiasing:
\begin{align}\label{eq19}
    \hat{\mathbb{P}}\left(c_i\mid d,\mathcal{D},\mathbf{C}_g \right) = \frac{\mathbb{P}_{\mathrm{obs}}\left(d_i\mid g,\tilde{\pi},\tilde{\lambda},\mathcal{D}, \mathbf{C}_g\right)}{\hat{\mathbb{P}}\left(d_i\mid \mathcal{D}\right)}
\end{align}

\section{Experiments}\label{sec6}
In this section, we aim to answer the following research questions (RQ) by experimental study:
\begin{itemize}
  \item {RQ1: Accuracy Performance}. Can our proposed methods outperform other talent recommendation baselines in terms of recommendation accuracy?
  \item {RQ2: Ablation Study}. Are the major components of our proposed model effective for improving recommendation accuracy?
  \item {RQ3: Bias Evaluation}. Can our proposed probability-based debiasing method effectively alleviate the bias issue?
\end{itemize}

\subsection{Experiment setup}\label{sec6.1}
\textbf{Dataset.} To evaluate the performance of our proposed L3TR model, we use two datasets: a private dataset collected from a human resource technology platform and a public  dataset. 
\begin{itemize}
  \item \textbf{HRT}: a dataset provided by a human resource service company and collected from their human resource technology (HRT) platform. The platform's main service is to recommend candidates for open positions and submit them for recruiters’ review. The time span of this data is around one year. The available information includes details of job postings, resumes, and positive matching pairs of job postings with resumes approved by the recruiters. We use this dataset to perform the talent recommendation task.
  \item \textbf{JobRec}: a public dataset\footnote{\url{https://www.kaggle.com/competitions/job-recommendation}} in a job recommendation challenge hosted by CareerBuilder in 2012. It consists of seven subsets, with each subset containing job application information spanning 13 days. We randomly select one of the subsets for the experiment. For each user, we have their demographic and professional information, alongside their work history, limited to job titles. 

  Although our proposed method is developed for talent recommendation, it can also be adapted for recommending job positions to job seekers, i.e., the job recommendation task. Thus, we use this dataset to test our proposed method for job recommendation.

\end{itemize}
Table \ref{table2} summarizes the statistics of the two datasets. The average number of tokens is calculated by the tokenizer of ChatGLM3-6B~\citep{zengglm}. The number of records means the number of matching pairs in the dataset. Examples of these two datasets are provided in Appendix~\ref{appendix_b}.

\begin{table}[h]
  \centering
\caption{Dataset Statistics.}\label{table2}
\begin{tabular}{l|ccccc}
\toprule
Dataset & \# Resumes & \# Tokens in Resume &\# Jobs & \# Tokens in Job Posting  & \# Records \\
\midrule
HRT & 29057  & 1214 & 1198  & 352 & 30810 \\
JobRec & 4821 & 44 & 10642 & 402  & 14868 \\
\bottomrule
\end{tabular}
\end{table}

\textbf{Baselines.} We compare our proposed models against two sets of baseline models. The first set consists of traditional deep learning models designed for the recommendation task, which we refer to as domain-specific models. We implement $4$ typical models as shown below:
\begin{itemize}
  \item {\bf PJFNN}~\citep{DBLP:journals/tmis/ZhuZXMXDL18} is a person-job fit model which leverages a convolutional neural network to extract semantic representations of job postings and resumes and matches them based on their cosine similarity.
  \item {\bf APJFNN}~\citep{DBLP:conf/sigir/QinZXZJCX18} carefully designs hierarchical recurrent neural networks to dynamically encode the requirement section of job postings and job/project experience section in resumes.
  \item {\bf BPJFNN}~\citep{DBLP:conf/sigir/QinZXZJCX18} is a simplified version of APJFNN, which utilizes BiLSTM to learn representations of each job posting and resume.
  \item {\bf TRBERT} is a twin tower model using pre-trained BERT~\citep{DBLP:conf/naacl/DevlinCLT19} as a text encoder and a multi-layer perceptron for predictions. We design and implement this model for talent (or job) recommendation task to evaluate the performance of domain-specific models based on pre-trained language models. 
\end{itemize}
The second set includes LLM-based models. We implement $7$ representative LLM-based baselines, selected along two dimensions: (1) whether fine-tuning is applied, and (2) whether the recommendation is performed through implicit or explicit methods. For non–fine-tuned models, we select open-source LLMs to implement implicit recommendation methods, and closed-source proprietary LLMs for explicit methods. Since open-source models are typically smaller while proprietary ones are considerably larger, this setup allows us to compare the performance of LLMs across different scales. In addition, we also consider large reasoning models with deep reasoning capability, which have been fine-tuned on various retrieval tasks. These models are expected to exhibit stronger logical inference and ranking abilities, making them suitable for comparison in our listwise recommendation setting.
\begin{itemize}
 \item {\bf Point-Imp} and {\bf List-Imp}, based on open-source LLMs, leverage the in-context learning method and follow the implicit method defined in Definition \ref{def3} for pointwise and listwise recommendations, respectively. 
 \item {\bf Point-Exp} and {\bf List-Exp} use the explicit method described in Definition \ref{def2} to prompt LLMs for both pointwise and listwise recommendations. A large commercial model, GLM-4~\citep{glm2024chatglm} (larger than 100B), is used in these baselines.
  \item {\bf TALLRec}~\citep{DBLP:conf/recsys/BaoZZWF023} is originally proposed for pointwise recommendations, which fine-tunes open-source LLMs to predict user preference over target items as a binary classification problem. TALLRec makes predictions by generating ``Yes'' or ``No'', making it difficult to perform the ranking needed in our task. To this end, we modify the output strategy to the implicit method, enabling it to produce matching scores for ranking. 
  \item {\bf Rank1}~\citep{weller2025rank1} is a pointwise ranking model based on reasoning LLMs. It leverages a large reasoning model to generate reasoning traces for query–document pairs, and distills this knowledge into a smaller reranker model. We use the 7B version for comparison, as this pointwise and reasoning-based model exhibits high latency during inference.
  \item {\bf ReasonRank}~\citep{liu2025reasonrank} is a listwise reasoning-based ranking model that adopts a two-stage training framework to enhance reasoning ability. It achieves state-of-the-art performance on reasoning-based retrieval benchmarks. We use both the 7B and 32B versions for comparison.
\end{itemize}
For LLM-based recommendation methods based on open-source LLMs, including our proposed method, we choose two LLMs of different sizes: ChatGLM3-6B~\citep{zengglm} and InternLM2.5-20B~\citep{cai2024internlm2}. Specifically, we use ChatGLM3-6B for experiments conducted on dataset HRT and InternLM2.5-20B on dataset JobRec, so that LLMs of different sizes are covered in our evaluation.

\textbf{Implementation Details.} To evaluate the recommendation performance, we adopt three widely-used metrics: Normalized Discounted Cumulative Gain (ND@$\{1,5\}$), Recall (R@$\{5,10\}$), and Mean Reciprocal Rank (MRR). The detailed computation procedures for each metric are provided in Appendix \ref{appendix_d}.

For the HRT dataset (talent recommendation), we randomly select 85\% of the job postings from $\mathcal{G}$ to form the training set $\mathcal{G}_{\text{train}}$, and use the remaining 15\% as the test set $\mathcal{G}_{\text{test}}$. The matching records whose job postings belong to $\mathcal{G}_{\text{train}}$ constitute $\mathcal{M}_{\text{train}}$, while those associated with $\mathcal{G}_{\text{test}}$ form $\mathcal{M}_{\text{test}}$. For the pointwise method, we perform a 1:1 negative sampling based on the records in $\mathcal{M}_{\text{train}}$. For the listwise method, we set the number of candidates per training instance to $N_{\text{train}}=4$. During testing, all models are evaluated with $N=15$ candidates per job posting. Experimental results with other values of 
$N_{\text{train}}$ and $N$ are provided in Appendix \ref{appendix_e}. Each test sample is constructed from a positive matching record in $\mathcal{M}_{\text{test}}$, where the candidate set consists of one positive resume and 14 randomly sampled negative resumes. The set of ID tokens is $\mathcal{D} = \{\text{``A''},\text{``B''},\dots,\text{``O''}\}$. In addition, we randomly sample 100 matching records from $\mathcal{M}_{\text{train}}$ to construct a subset $\mathcal{T}$, which is used to evaluate the bias issue and estimate the global prior bias. This step is only applied to listwise methods that require debiasing. For the JobRec dataset (job recommendation), we adopt a similar procedure. The only difference lies in the splitting order: we first divide the resume set $\mathcal{R}$ into training and test subsets, then partition the matching records $\mathcal{M}$ accordingly.

The domain-specific models are implemented based on RecBole~\citep{DBLP:conf/cikm/ZhaoHPYZLZBTSCX22}. Rank1 and ReasonRank are implemented based on their public models. For methods involved with large commercial models (Point-Exp and List-Exp), we call the API of GLM-4-Plus provided by ZhipuAI\footnote{\url{https://bigmodel.cn/dev/api/normal-model/glm-4}}. TALLRec and our proposed methods share the same open-source LLMs. We adopt the LoRA~\citep{DBLP:conf/iclr/HuSWALWWC22} method with the rank of 8 to fine-tune the models on the training set for 2 epochs. The number of special tokens for each resume (or job posting in the JobRec dataset) is set as $S=10$. Experimental results with other values of 
$S$ are provided in Appendix \ref{appendix_e}. The prompt templates for LLM-based methods are provided in Appendix~\ref{appendix_c}.

\subsection{Performance Comparison}

\begin{table}[t]
  \caption{Recommendation Accuracy on HRT.}
  \label{table3}
  \centering
    \begin{tabular}{l|l|rrrrr}
    \toprule
   Type & {Model} & ND@5  &  ND@10 &  R@1 & R@5 &  MRR \\
    \midrule
   \multirow{4}{3cm}{Domain-specific Models} & PJFNN &  .3430 & .4236 & .1551 & .5248 & .3348 \\
   & BPJFNN & .5077 & .5645 & .2607  & .7409 & .4614  \\
   & APJFNN & .3872 & .4857 & .1683 &  .6056 & .3646 \\
   & TRBERT & .6078 & .6604 & .3927 & .7937 & .5716\\
    \midrule
   \multirow{8}{3cm}{LLM-based Models} & List-Imp (6B) & .2087 & .3062 & .0792 & .3564 & .2276\\
   & List-Exp ($>$100B) & .7982 & .8166 & .6815 & .8311 & .7781\\
   & Point-Imp (6B) & .5252 & .5973 & .3267 & .7030 & .5033 \\ 
   & Point-Exp ($>$100B) & .7850 & .7981 & .6238 & .8191 & .7484 \\
   & TALLRec (6B) & .8442 & .8532 & 7046 & .8838 & .8134    \\
   & Rank1 (7B) & .7833 & .7975 & .6485 & .8927 & .7577 \\
   & ReasonRank (7B) & .6643 & .6977 & .5307 & .7782 & .6491 \\
   & ReasonRank (32B) & .8046 & .8198 & .7104 & .8886 & .7879 \\   
    \midrule 
   \multirow{1}{3cm}{Our Models} & L3TR (6B) & {\bf .8518}& {\bf .8659} & {\bf .8102} & {\bf .8868} & {\bf .8531} \\
    \bottomrule
    \end{tabular}%
  \end{table}%
  
Table~\ref{table3} and \ref{table4} show the performance in terms of accuracy on HRT and JobRec, respectively.  Since the key information of each resume in JobRec only contains job titles, we omit APJFNN, which highly relies on rich textual content. L3TR is equipped with the debiasing method described in Eq. (\ref{eq19}). We will discuss more details about the bias issue in Section~\ref{sec6.4}.
  
\begin{table}[!htbp]
      \caption{Recommendation Accuracy on JobRec. }
      \label{table4}
      \centering
    \begin{tabular}{l|l|rrrrr}
    \toprule
   Type & {Model} & ND@5  &  ND@10 &  R@1 & R@5 &  MRR \\
    \midrule
    \multirow{4}{3cm}{Domain-specific Models} & PJFNN &  .2080 & .3132 & .0704 & .3527 & .2285 \\
    & BPJFNN & .2008 & .3051 & .0669  & .3406 & .2225  \\
    & TRBERT & .1967 & .2976 & .0663 & .3351 & .2197\\
    \midrule
     \multirow{8}{3cm}{LLM-based Models} & List-Imp (20B)  & .5970 & .6444 & .4620 & .7191 & .5860 \\
    & List-Exp ($>$100B) & .7539 & .7719 & .5994 & .8844 & .7234 \\
    & Point-Imp (20B) & .7119 & .7466 & .5386 & .8616 & .6790 \\ 
    & Point-Exp ($>$100B) & .5800 & .6288 & .3627 & .7733 & .5417 \\
    & TALLRec (20B) & .9003 & .9125 & .8437 & .9483 & .9019 \\
    & Rank1 (7B) & .6559 & .6969 & .4496 & .8420 & .6142 \\
    & ReasonRank (7B) & .6429 & .6842 & .4746 & .7940 & .6160 \\
    & ReasonRank (32B) & .6935 & .7195 & .5745 & .8123 & .6814 \\ 
    \midrule 
    \multirow{1}{3cm}{Our Models} & L3TR (20B) & {\bf .9227} & {\bf .9313} & {\bf .8794} & {\bf .9724} & {\bf .9231} \\
    \bottomrule
    \end{tabular}%
\end{table}%

First, on both datasets, L3TR outperforms all other baselines in terms of all accuracy performance measures, including methods based on large-scale LLMs. Our proposed model demonstrates a significant improvement in the top-ranked positions of the list. Compared to TALLRec, L3TR achieves a 13.11\% increase in R@1 on the HRT dataset and a 3.76\% increase on the JobRec dataset. 

Second, for implicit recommendation methods based on smaller-scale open-source LLMs (in our case, ChatGLM3-6B and InternLM2.5-20B), the pointwise approach (Point-Imp) performs better than its corresponding listwise version (List-Imp), which means that LLMs can understand a pair of a job posting and a resume better but struggle to handle an entire candidate set. This also confirms that LLM-based listwise recommendation remains challenging even with a longer context window. On the other hand, explicit recommendation methods based on large-scale closed-source LLMs possess stronger contextual understanding capabilities, allowing them to utilize the listwise setting to compare information across multiple resumes and thereby achieve better recommendation results than their pointwise counterparts. Note that pointwise methods generally consume more tokens overall, since job posting (or resume in job recommendation task) information needs to be processed repeatedly $N$ times.

Third, although large reasoning models have been trained on large-scale retrieval data for retrieval and ranking task, they still performs worse than TALLRec in most cases, which is specifically fine-tuned on recommendation data. This suggests the necessity of fine-tuning to align LLMs with the recommendation task. Fourth, most of domain-specific models (PJFNN, BPJFNN, and APJFNN) get limited performance in the top place. Their low results on R@1 indicate their limited capacity to comprehend text information compared with LLMs.

\subsection{Ablation Study}\label{sec6.3}
To verify the effectiveness of each part in L3TR, we implement several variants of L3TR for an ablation study:
\begin{itemize}
  \item {\bf $\text{L3TR}^{-}$}. We remove all designs in Figure~\ref{fig5} and fine-tune the LLM with the listwise data using the implicit method.
  \item {\bf L3TR-B}. We remove the \uline{b}lock attention mechanism. 
  \item {\bf L3TR-I}. We remove the \uline{I}D sampling procedure in Eq (\ref{eq4}).
  \item {\bf L3TR-L}. We remove the \uline{l}ocal position encoding method.
\end{itemize}

For a fair comparison, the results of L3TR reported in this section are obtained without any additional debiasing methods. The evaluation results are shown in Table~\ref{table5} and Table~\ref{table6} for the two datasets, respectively. The full L3TR model achieves the best performance under all metrics, confirming that the three components jointly contribute to the overall gains. Among them, block attention is the most critical design, as removing it (L3TR-B) leads to the largest performance drop. In contrast, removing random ID assignment (L3TR-I) or local positional encoding (L3TR-L) results in more moderate declines.

Besides, we observe that $\text{L3TR}^{-}$ significantly outperforms List-Imp on both datasets, verifying that it's effective to fine-tune LLMs using implicit methods on listwise datasets. However, the performance of $\text{L3TR}^{-}$ still lags behind the pointwise fine-tuning method, TALLRec. A possible reason for this gap is that the candidate set length during training ($N_\text{train}= 4$ ) is smaller than that in the test data ($N = 15$). Given resource constraints, the candidate set length in the training data is always difficult to fully scale with that in the test data. Therefore, it is crucial to design a more effective listwise fine-tuning method that can overcome the discrepancy between $N_\text{train}$ and $N$.

\begin{table}[!htbp]
  \caption{Ablation Results on HRT. }
  \label{table5}
  \centering
    \begin{tabular}{l|C{1.8cm}C{2cm}C{1.8cm}|rrrrr}
    \toprule
    {Model} & Block Attention & Random ID Assignment & Local Positional Encoding& ND@5  &  ND@10 &  R@1 & R@5 &  MRR \\
    \midrule
    $\text{L3TR}^{-}$  & $\times$ & $\times$ & $\times$ & .7692 & .7999 & .7046 & .8284 & .7682 \\
    L3TR-B & $\times$ & $\surd$ & $\surd$ & .6248 & .6650 & .5248 & .7063 & .6260 \\
    L3TR-I & $\surd$ & $\times$ & $\surd$ & .8156 & .8501 & .7721 & .8516 & .8195 \\
    L3TR-L & $\surd$ & $\surd$ & $\times$ & .8204 & .8419 & .7640 & .8696 & .8177 \\
    \midrule 
    L3TR & $\surd$ & $\surd$ & $\surd$ &   {\bf .8379}& {\bf .8545} & {\bf .7953} & {\bf .8762} & {\bf .8377} \\
    \bottomrule
    \end{tabular}%
\end{table}%
  
  \begin{table}[t]
  \caption{Ablation Results on JobRec. }
  \label{table6}
  \centering
    \begin{tabular}{l|C{1.8cm}C{2cm}C{1.8cm}|rrrrr}
    \toprule
    {Model} & Block Attention & Random ID Assignment & Local Positional Encoding& ND@5  &  ND@10 &  R@1 & R@5 &  MRR \\
    \midrule
    $\text{L3TR}^{-}$  & $\times$ & $\times$ & $\times$ & .8894 & .9022 & .8295& .9351 & .8812 \\
    L3TR-B & $\times$ & $\surd$ & $\surd$ & .8157 & .8576 & .7889 & .8358 & .8232 \\
    L3TR-I & $\surd$ & $\times$ & $\surd$ & .9023 & .9097 & .8581 & .9402 & .9017 \\
    L3TR-L & $\surd$ & $\surd$ & $\times$ & .9031 & .9125 & .8665 & .9410 & .9065 \\
    \midrule 
    L3TR & $\surd$ & $\surd$ & $\surd$ &  {\bf .9124} & {\bf .9208} &  {\bf .8683} & {\bf .9638} & {\bf .9167} \\
\bottomrule
\end{tabular}%
\end{table}%

\subsection{Bias Evaluation Results}\label{sec6.4}
In this subsection, we focus on answering the third question (RQ3) by evaluating the effectiveness of listwise methods in terms of bias mitigation. We examine four listwise recommendation models: List-Imp, List-Exp, L3TR$^-$, and L3TR. For models based on open-source LLMs, we compare their variants with different debiasing methods. Among the three commonly used accuracy metrics, MRR is a global indicator that does not depend on a predefined top-$K$ cutoff. Therefore, we choose this metric for our analysis in this section.

\begin{figure}[!htbp]
    \centering
    \includegraphics[width=0.5\linewidth]{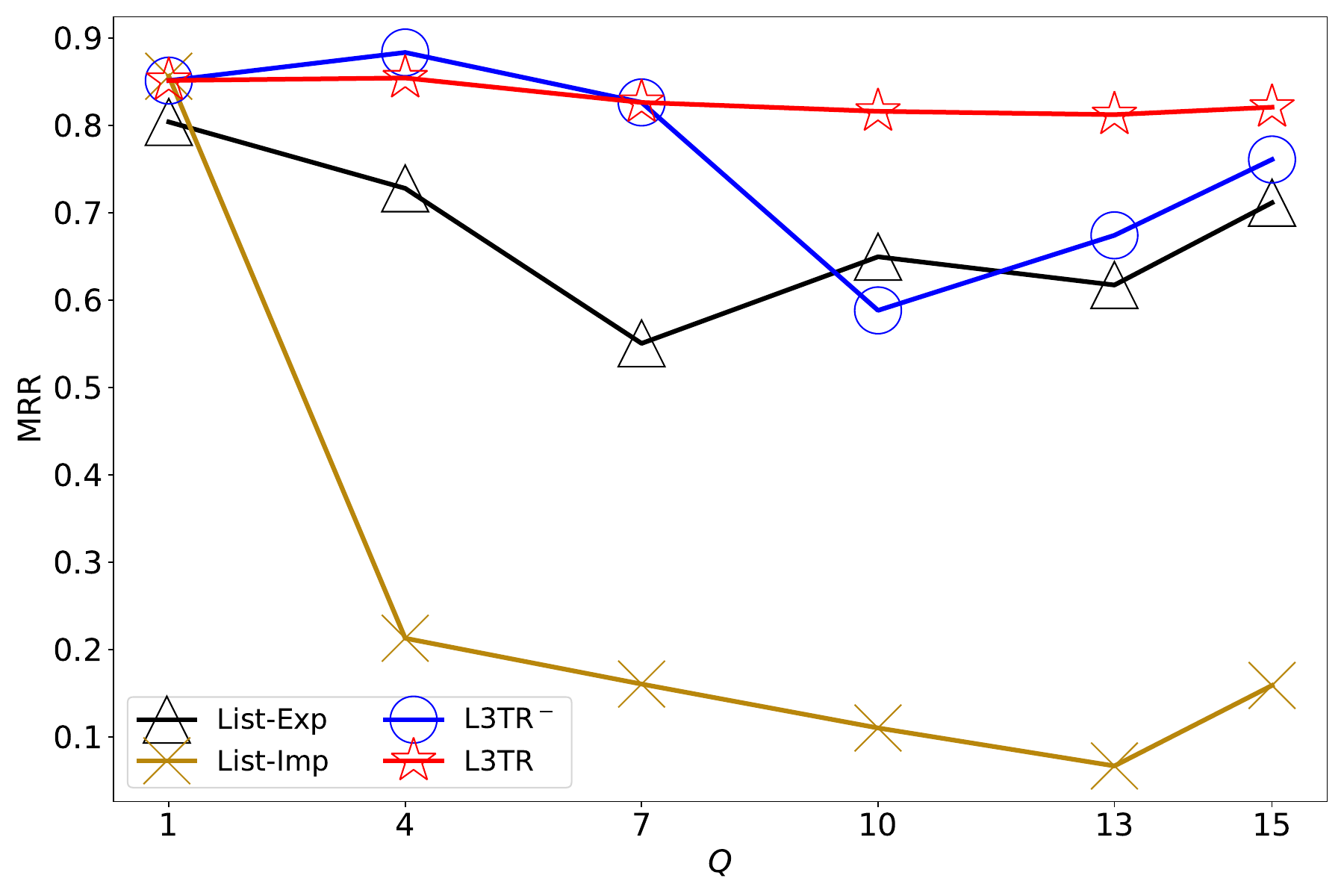}
    \caption{Accuracy Fluctuations with Different $Q$.}
    \label{fig7}
\end{figure}

We focus on the HRT dataset, set the number of candidates to $N=15$, and put the matched resume at the $Q$-th position. For each value of $Q$, we randomly construct five different candidate sets and report the average evaluation results. Figure \ref{fig7} illustrates the accuracy fluctuations of various models. 

From Figure \ref{fig7}, we can see that the untuned small LLM (List-Imp with ChatGLM3-6B) exhibits severe bias. When the matched resume appears in the first position of the candidate list ($Q=1$), it performs well. However, as $Q$ increases, its accuracy drops sharply. When the matched resume is placed at the end of the candidate list ($Q=15$), performance recovers slightly but remains significantly lower than the result at $Q=1$. Although large commercial LLM (List-Exp with GLM-4-Plus) performs better overall than List-Imp, it still exhibits notable performance fluctuations. Its performance continues to exhibit a pattern of being higher at both ends and lower in the middle. This is similar to the findings reported by \citet{liu2024lost} in RAG, where key documents placed in the middle of all retrieved documents are more likely to be overlooked by LLMs. Fine-tuning LLMs with domain-specific data can significantly enhance their performance. However, L3TR$^-$, which utilizes listwise data and the LoRA method to fine-tune the LLM without special designs, remains highly sensitive to the position of the matched resume. In contrast, L3TR incorporates local position encoding and block attention, enabling it to better capture the relationships between the job description and each resume. As a result, L3TR achieves highly stable performance that is almost unaffected by changes in $Q$.

Furthermore, we decompose the observed bias of L3TR and L3TR$^-$. Based on the analysis in Section~\ref{sec5.1}, we use Eq. (\ref{eq13}) to approximate the LLM’s preference for content located at different positions. We fix the matched resume at position $Q$ and randomly shuffle the ID tokens of the candidate resumes in each list so that they are no longer in alphabetical order. In this way, we remove the effect of the ID token (token bias) on the performance so that we can measure the position bias of different models. We repeat the evaluation five times and report the average result. In addition, we record both the performance range and the standard deviation of each model for different $Q$. The results are shown in Figure~\ref{fig8} and Table~\ref{tab6}. 

\vspace{0.3cm}
\begin{minipage}{0.95\textwidth}
  \begin{minipage}[T]{0.5\textwidth}
    \centering
    \includegraphics[width=\linewidth]{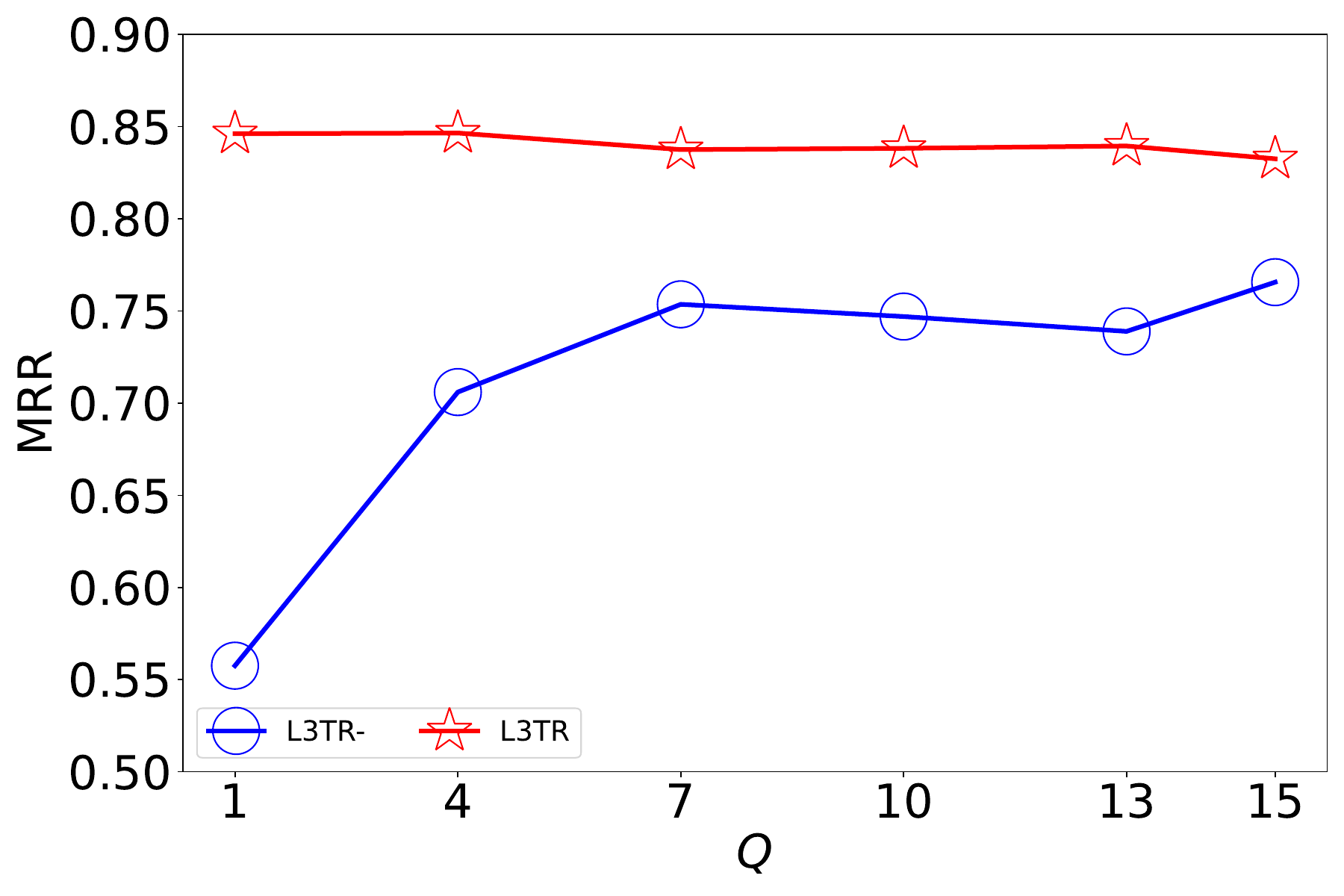} 
    \captionof{figure}{Position Bias.}
    \label{fig8}
  \end{minipage}
  \hfill
  \begin{minipage}[T]{0.47\textwidth}
    \centering
    \captionof{table}{Performance Range and the Standard Deviation}
    \label{tab6}
        \begin{tabular}{l|rr}
      \toprule
     {Model} & Range & Std \\
      \midrule
      L3TR$^-$ &  0.2081 & 0.0780 \\
      L3TR & 0.0140 & 0.0054 \\
      \bottomrule
    \end{tabular}
    \end{minipage}
\end{minipage}
\vspace{0.3cm}

Consistent with the results shown in Figure \ref{fig7}, L3TR exhibits highly stable performance with minimal position bias. In contrast, L3TR$^-$ shows a different pattern from Figure \ref{fig7}: it has a lower preference for content at position 1. This suggests that the strong performance of L3TR$^-$ at $Q = 1$ in Figure \ref{fig7} can largely be attributed to the model’s bias toward the ID token ``A'', since during the evaluation in Figure \ref{fig7}, the resume at the first position was always assigned the ID token ``A''. Moreover, the range and standard deviation shown in Table \ref{tab6} clearly demonstrate that directly fine-tuning an LLM with listwise data can cause severe position bias issues, whereas our designs effectively mitigate this problem.

\begin{figure}[!htbp]
  \centering
  \subfigure[]{\label{fig9a}
  \includegraphics[width=0.44\linewidth]{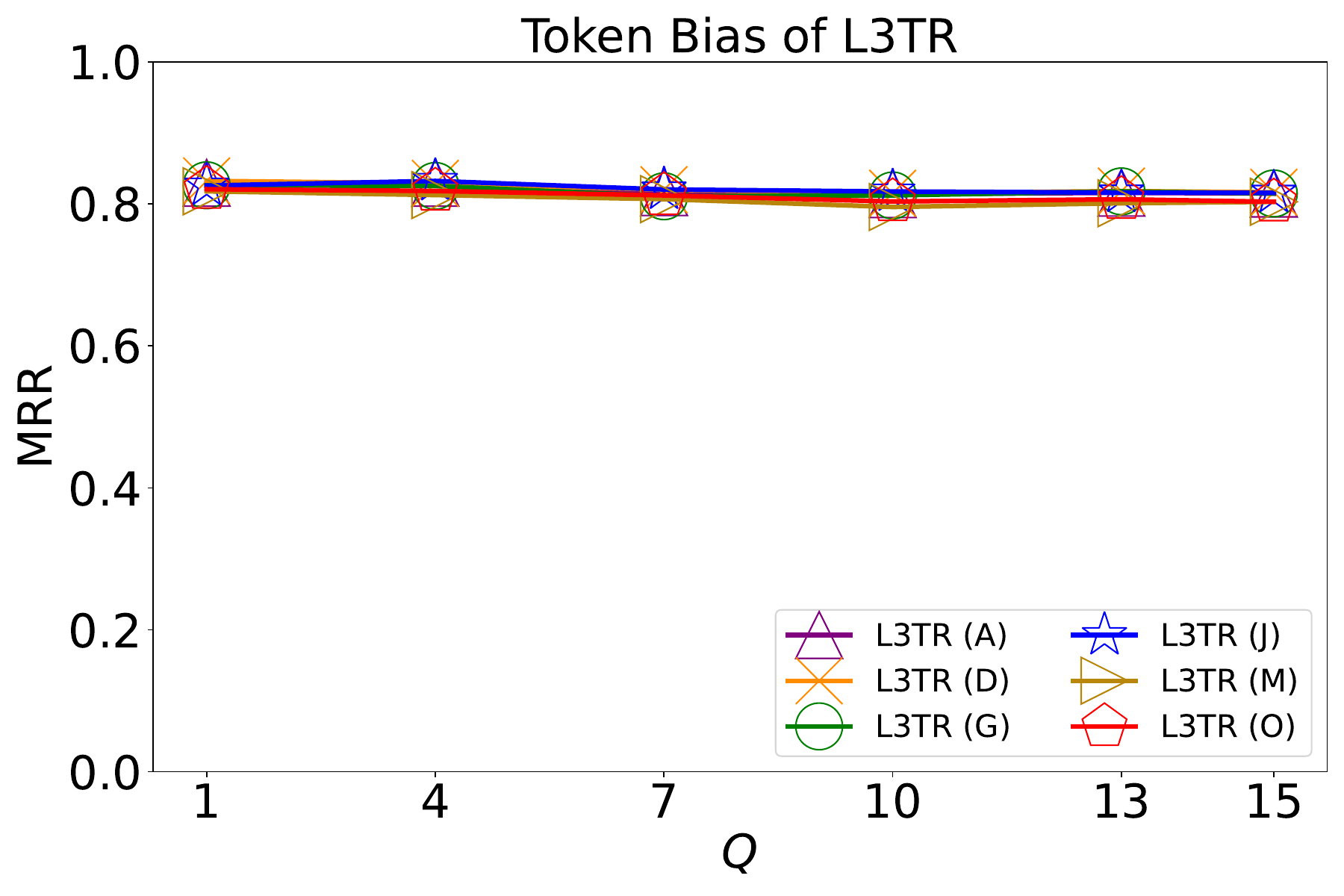}
  }
  \subfigure[]{\label{fig9b}
  \includegraphics[width=0.44\linewidth]{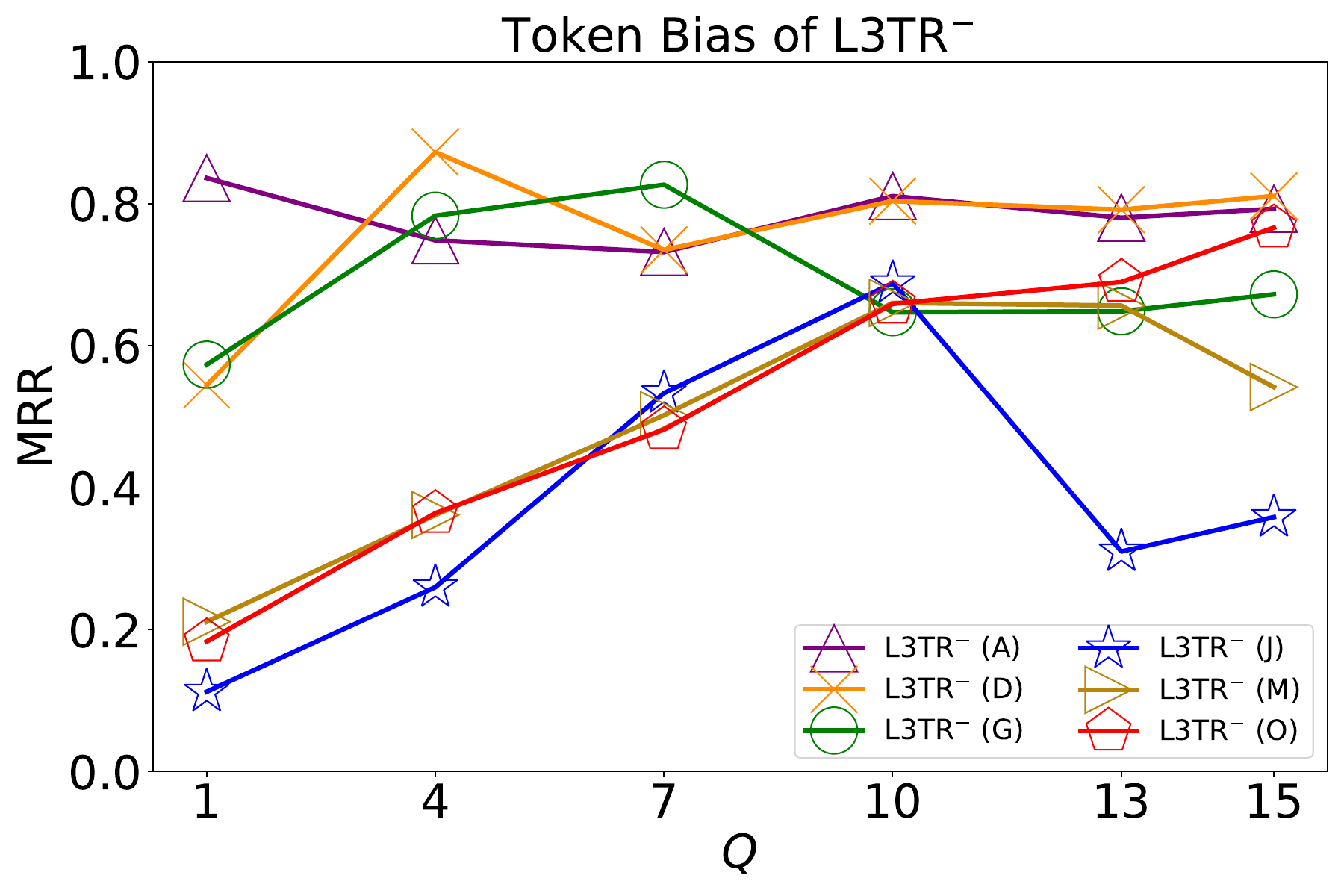}
 }
  \caption{Token Bias at Different Positions of L3TR and L3TR$^-$.}
  \label{fig9}
\end{figure}

\begin{figure}[!htbp]
  \centering
  \subfigure[]{\label{fig10a}
  \includegraphics[width=0.44\linewidth]{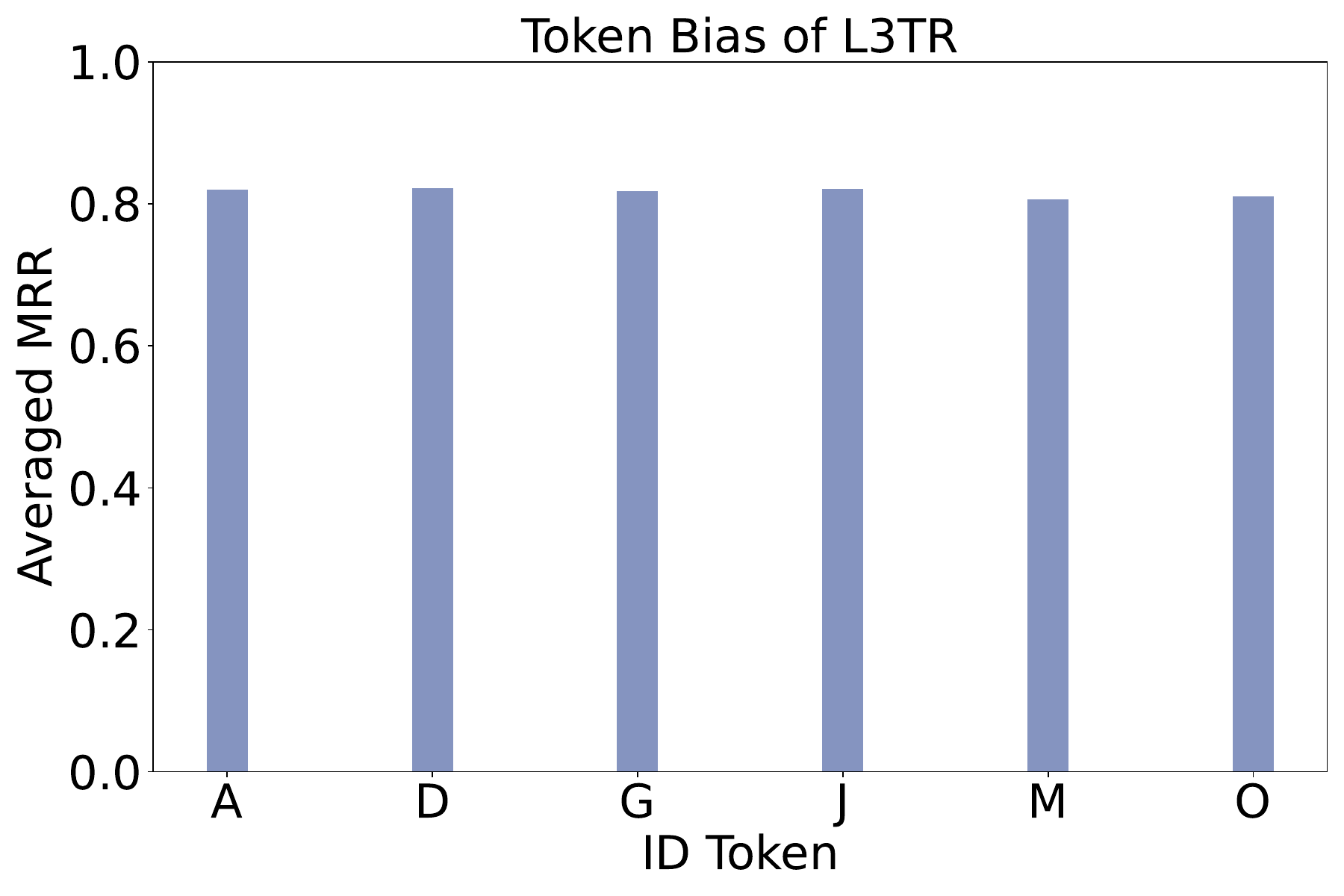}
  }
  \subfigure[]{\label{fig10b}
  \includegraphics[width=0.44\linewidth]{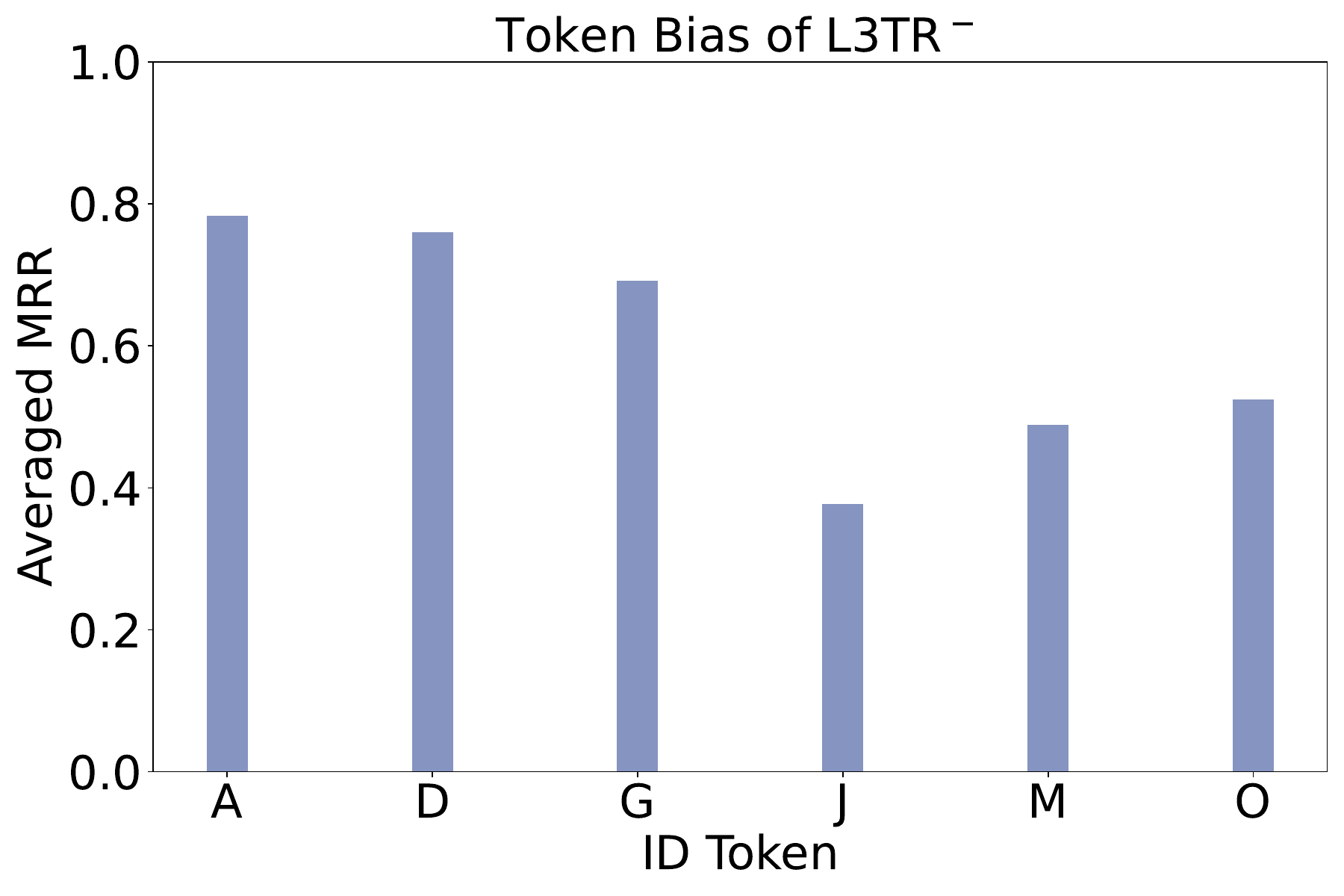}
 }
  \caption{Token Bias of L3TR and L3TR$^-$.}
  \label{fig10}
\end{figure}

For token bias, we select a set of ID tokens from $\mathcal{D} = \{\text{``A''}, \text{``B''}, \cdots, \text{``O''}\}$, the 15 tokens used to identify each candidate. Based on considerations of both representativeness and efficiency, we choose six of them: {``A'', ``D'', ``G'', ``J'', ``M'', ``O''} in this experiment. For each job, we first put its matched resume at position $Q$ and then iterate through each chosen token, assigning it to that position $Q$. For each ID token, we evaluate the model’s accuracy for this token at the position $Q$ according to Eq. (\ref{eq10}), which is shown in Figures \ref{fig9a} and \ref{fig9b} for L3TR and L3TR$^-$, respectively. We then average the results across all positions to obtain the model’s overall preference for different ID tokens, as presented in Figure \ref{fig10}.

Figure \ref{fig9a} indicates that placing the same ID at different positions has minimal influence on L3TR's recommendation performance, which further confirms that L3TR exhibits only mild position bias. Although L3TR’s preference varies slightly across different tokens at each position, such variation remains limited. In contrast, both the position and the ID token have a more significant impact on L3TR$^-$'s performance. As illustrated in Figure \ref{fig9b}, L3TR$^-$ exhibits distinct token preferences depending on position. For example, it strongly favors token ``A'' in the first position but clearly prefers token ``D'' in the fourth. Interestingly, the position yielding the best performance for each token aligns with its alphabetical order (e.g., ``A'' at position 1, ``D'' at 4, and ``G'' at 7), suggesting that L3TR$^-$ tends to form spurious correlations between token identity and positional index. A comparison between Figures~\ref{fig10a} and~\ref{fig10b} further confirms that both models show token-specific preferences, but the variations are far more pronounced for L3TR$^-$. Based on Figures~\ref{fig7}, \ref{fig8}, \ref{fig9b}, and \ref{fig10b}, these results indicate that the strong performance of L3TR$^-$ when the matched resume appears in the first position of the candidate list ($Q=1$) mainly stems from its bias toward the ID token ``A'', rather than an inherent preference for the first position itself.

To leverage the bias patterns observed in Figure \ref{fig7}, we employ the pre-ranker introduced in Section \ref{sec5.2} to pre-sort the candidate list, thereby increasing the likelihood that the matching resume appears near the front. The pre-ranker is applicable to both open-source and closed-source models. In addition, we adopt a probability-based debiasing method to mitigate both token bias and position bias. All results are presented in Table \ref{table7}.  

The pre-ranker method pushes potentially matching resumes toward the front of the candidate list, as LLMs typically favor items placed at the top positions. With this adjustment, List-Exp, List-Imp, and L3TR$^-$ achieve MRR improvements ranging from 1.14\%–69.77\%. For models heavily affected by bias, such as List-Imp, the pre-ranker yields gains of over 50\%. Meanwhile, the probability-based debiasing method described in Eq. (\ref{eq19}) also benefits various open-source LLM-based models. For instance, all metrics of L3TR$^-$ improve by at least 4\%. When combined, the two methods deliver even greater performance improvements. Although the observed bias in L3TR is relatively mild, it still gains from both the pre-ranker and probability-based debiasing approaches.

In summary, both the pre-ranker and the probability-based debiasing method enhance model performance beyond the models’ intrinsic capabilities. Models exhibiting stronger bias benefit more substantially, while those with lighter bias still achieve meaningful improvements.

\begin{table}[!htbp]
      \caption{Evaluations of Different Debiasing Methods on HRT.}
      \label{table7}
      \centering
    \begin{tabular}{l|rrrrr}
    \toprule
    {Model} & ND@5  &  ND@10 &  R@1 & R@5 &  MRR \\
    \midrule
    List-Exp & .7982 & .8166 & .6815 & .8311 & .7781 \\
    + Pre-ranker & .7989 & .8195 & .6997 & .8729 & .7870 \\
    \midrule
    List-Imp &  .2087 & .3062 & .0792 & .3564 & .2276\\
    + Pre-ranker & .3946  & .4750  & .2277  &  .5660 & .3864  \\
    + Debiasing Method & .2165 & .3102 & .0859 & .3616 & .2303 \\
    + Pre-ranker and Debiasing Method  & .3972  & .4788 & .2524 & .5437 & .3975 \\
    \midrule
    L3TR$^-$ & .7692 & .7999 & .7046 & .8284 & .7682 \\
    + Pre-ranker & .8091 & .8348 & .7525  & .8663 & .8054  \\
    + Debiasing Method & .8298 & .8458 & .7756 & .8812 & .8245 \\
    + Pre-ranker and Debiasing Method  & .8336 & .8493 & .7904 & .8815 & .8303 \\
    \midrule
    L3TR  &{ .8379}& { .8545} & { .7953} & { .8762} & { .8377}  \\
    + Pre-ranker & .8389 & .8552 & .7970  & .8779 & .8381  \\
    + Debiasing Method & .8445 & .8651 & .8027 & .8795 & .8422 \\
    + Pre-ranker and Debiasing Method  &  {\bf .8518}& {\bf .8659} & {\bf .8102} & {\bf .8868} & {\bf .8531} \\
    \bottomrule
    \end{tabular}%
\end{table}%

\section{Conclusion}\label{sec7}
In this paper, we introduced L3TR, a novel framework for aligning large language models with the listwise talent recommendation task. Unlike conventional pointwise approaches, for each job, L3TR directly processes a set of resumes and outputs a ranked candidate list, avoiding redundant processing of each resume. We proposed an implicit recommendation strategy to finetune LLMs and introduce an ID sampling method to resolve the inconsistency between candidate set sizes in the training phase and the inference phase. Then we design a block attention mechanism together with a local positional encoding method to enhance inter-document relationship understanding and mitigate position issue. We further provide a theoretical analysis to validate the design of the local positional encoding method. In addition, we decomposed the observed bias patterns into position bias and token bias, and developed evaluation methods to detect biases as well as training-free debiasing methods. We conduct extensive experiments on two real-world datasets to study the position and token bias and evaluate the performance of our proposed framework and methods. Results demonstrate that L3TR not only improves recommendation accuracy, but also alleviates position bias and token bias.

Our findings highlight both the potential and the challenges of applying LLMs to listwise recommendation. LLMs can leverage their strong language understanding ability to model complex job–resume relationships and improve recommendation accuracy. However, their vulnerability to bias underscores the need for adaptation. The study of position and token bias not only advances our understanding of LLM behaviors in ranking settings but also provides a foundation for developing more robust and interpretable recommendation systems.

Several research directions remain open for exploration. First, our proposed framework has the potential to be applied to other recommendation scenarios, where both the recommender (e.g., user) and recommendee (item) are described by a lengthy set of tokens. Future work includes applying our method to a broader range of scenarios as demand arises and the relevant data become available. Second, fairness represents another critical consideration in talent recommendation. This issue warrants thorough investigation. Advancing in these directions will pave the way for building more effective LLM-based recommendation systems in talent recruitment service and beyond.

\bibliographystyle{ACM-Reference-Format}
\bibliography{ref}

\appendix
\section*{Appendix} 

\section{Notation}\label{appendix_notation}

\begin{longtable}{
>{\raggedright\arraybackslash}p{3.2cm}
>{\raggedright\arraybackslash}p{0.74\textwidth}
}
\caption{Notation and Symbol Definitions in the Main Text.}
\label{tab:notation_main}
\\
\toprule
\textbf{Symbol} & \textbf{Meaning} \\
\midrule
\endfirsthead

\toprule
\textbf{Symbol} & \textbf{Meaning} \\
\midrule
\endhead

\midrule
\multicolumn{2}{r}{\emph{Continued on next page}} \\
\endfoot

\bottomrule
\endlastfoot

\multicolumn{2}{l}{\textbf{Problem formulation and ranking}}\\
$\mathcal{G}$ & Set of job postings. \\
$\mathcal{R}$ & Set of resumes (talent pool). \\
$g_i$ & The $i$-th job posting, $g_i \in \mathcal{G}$. \\
$r_j$ & The $j$-th resume, $r_j \in \mathcal{R}$. \\
$\mathcal{M}$ & Historical matching records set, $\mathcal{M}=\{(g_i,r_j,y_{g_i,r_j})\}$. \\
$y_{g_i,r_j}$ & Ground-truth match label between $g_i$ and $r_j$; $y_{g_i,r_j}\in\{0,1\}$. \\
$\mathcal{C}_g$ & Candidate resume set for job $g$, $\mathcal{C}_g=\{c_1^g,c_2^g,\dots,c_N^g\}$. \\
$\underline{\mathcal{C}}_g$ & Ranked candidate list for job $g$. \\
$c_j^g$ & The $j$-th candidate resume in $\mathcal{C}_g$, $c_j^g\in R$. \\
$N$ & Number of candidates in a list (candidate list length). \\
$F$ & Recommender that outputs matching scores. \\
$\hat{y}_j^g$ & Predicted matching score for candidate $c_j^g$ under job $g$. \\
$\hat{\mathbf{y}}^{g}$ & Vector of predicted scores for all candidates: $\hat{\mathbf{y}}^{g}=(\hat{y}_1^g,\dots,\hat{y}_N^g)$. \\
$\tau_g$ & Ranking permutation induced by sorting predicted scores (descending). \\
$\tau_g(\ell)$ & Index of the candidate in $C_g$ ranked at position $\ell$. \\
$\langle \cdots \rangle$ & Ordered list notation (ranked candidate list). \\

\addlinespace
\multicolumn{2}{l}{\textbf{Prompt construction and LLM outputs}}\\
$x^g$ & Prompt text for job $g$ after formatting job posting and candidate resumes. \\
$x_i^g$ & The $i$-th token in prompt $x^g$. \\
$n_g$ & Number of tokens in prompt $x^g$. \\
$\texttt{BuildPrompt}(\cdot)$ & Prompt generator function. \\
$\texttt{assign}(\mathcal{C}_g)$ & ID assignment operator mapping candidates in $C_g$ to distinct identifiers used in the prompt/output. \\
$\texttt{ID}(c_j^g)$ & Assigned ID token for candidate resume $c_j^g$. \\
$o^g$ & LLM generated output token sequence for job $g$. \\
$o_t^g$ & The $t$-th output token in $o^g$. \\
$|o^g|$ & Output length (number of generated tokens). \\
$F_{\mathrm{LLM}}(\cdot)$ & LLM forward/generation function applied to prompt. \\
$\mathbb{P}(\texttt{ID}(c_j^g))$ & Probability that the LLM generates the ID token of $c_j^g$ as the \emph{first} output token. \\
$\hat{y}_j^g$ & Normalized first-token probability score:
$\hat{y}_j^g=\frac{\mathbb{P}(\texttt{ID}(c_j^g))}{\sum_{s=1}^{N}\mathbb{P}(\texttt{ID}(c_s^g))}$. \\

\addlinespace
\multicolumn{2}{l}{\textbf{Training sample construction and loss}}\\
$\mathcal{R}_{g_i}^{+}$ & Positive resume set for job $g_i$ (matched resumes according to $M$). \\
$\mathcal{R}_{g_i}^{-}$ & Negative resume set for job $g_i$ (unmatched resumes according to $M$). \\
$N_{\mathrm{train}}$ & Number of resumes in the \emph{training} prompt (typically $N_{\mathrm{train}}<N$). \\
$\mathcal{C}_{g_i,r_k}$ & Candidate set constructed for training from positive record $(g_i,r_k,1)$: includes $r_k$ and $N_{\mathrm{train}}-1$ negatives. \\
$\mathcal{I}$ & A large ID pool from which training IDs are sampled. \\
$\texttt{sample}(\mathcal{I},N_{\mathrm{train}})$ & Operator sampling $N_{\mathrm{train}}$ unique IDs uniformly from $\mathcal{I}$. \\
$\mathcal{L}_{g_i,r_k}$ & Training loss for sample built from record $(g_i,r_k,1)$. \\
$y_{g_i,c_j}$ & Ground-truth label for candidate $c_j$ w.r.t.\ job $g_i$ in the constructed training list. \\

\addlinespace
\multicolumn{2}{l}{\textbf{Block attention (prompt blocks and special tokens)}}\\
$b_0$ & Number of tokens in the task-description block (block boundary). \\
$b_1$ & Block boundary after the job-posting block; job-posting length is $b_1-b_0$. \\
$b_j$ & Boundary index associated with the $j$-th resume block; resume $c_j$ spans $(b_j+1,\cdots,b_{j+1})$. \\
$S$ & Number of learnable special tokens inserted before each resume. \\
$x^{c_j}_s$ & The $s$-th special token (learnable) associated with resume $c_j$, $s\in\{1,\cdots,S\}$. \\
$x'$ & Modified prompt after inserting special tokens and applying index remapping. \\

\addlinespace
\multicolumn{2}{l}{\textbf{Local positional encoding (embeddings and attention notation)}}\\
$\texttt{emb}(\cdot)$ & Token embedding layer/function. \\
$\mathbf{x}_i$ & Token embedding vector at position $i$ (after embedding lookup); $\mathbf{x}_i\in\mathbb{R}^{d}$. \\
$d$ & Embedding dimension; also used as dimension parameter in positional encoding. \\
$\texttt{PE}(i,d)$ & Positional encoding function producing position embedding for position $i$. \\
$\boldsymbol{p}_i$ & Global positional embedding at position $i$: $\boldsymbol{p}_i=\texttt{PE}(i,d)$. \\
$\boldsymbol{q}_i,\boldsymbol{k}_i,\boldsymbol{v}_i$ & Query, key, value vectors at position $i$. \\
$f_q,f_k,f_v$ & Functions (linear projections / position embedding injection mechanisms) mapping $(\mathbf{x}_i,\boldsymbol{p}_i)$ to $\boldsymbol{q}_i,\boldsymbol{k}_i,\boldsymbol{v}_i$. \\
$\boldsymbol{p}'_i$ & \emph{Local} positional embedding for token $\mathbf{x}_i$ within resume block $c_j$:
$\boldsymbol{p}'_i=\texttt{PE}(i-b_j+b_1,d)$. \\

\addlinespace
\multicolumn{2}{l}{\textbf{Bias analysis: datasets, IDs, permutations}}\\
$\mathcal{T}$ & Test dataset constructed from a subset of matching records by forming candidate lists for evaluation. \\
$\mathcal{M}'$ & Subset of historical matching records $\mathcal{M}$ used to build $\mathcal{T}$. \\
$\mathbf{C}_g = <c_1^g,c_2^g,\cdots,c_N^g>$ & The ordered version of the candidate list $\mathcal{C}_g$ where candidates are alphabetically sorted by their names by default. \\
$Q$ & Position index (in the candidate list) where the matched resume is placed during controlled perturbation. \\
$\mathcal{D}=\{d_1,\cdots,d_N\}$ & Set of ID tokens assigned to an ordered candidate list (e.g., $\{\text{A}, \text{B}, \cdots \}$). \\
$d_i$ & A specific ID token in $\mathcal{D}$. \\
$\Lambda(\mathbf{C}_g)$ & Set of permutations of the candidate list $\mathbf{C}_g$. \\
$\lambda\in\Lambda(\mathbf{C}_g)$ & A permutation of candidates in $\mathbf{C}_g$ (reordering resumes). \\
$\Pi(\mathcal{D})$ & Set of permutations of the ID token set $\mathcal{D}$. \\
$\pi\in\Pi(\mathcal{D})$ & A permutation of ID assignments (reordering IDs). \\
$\mathcal{B}$ & Reference ID set used for token-bias estimation (distinct from $\mathcal{D}$). \\
$|\mathcal{B}|$ & Size of $\mathcal{B}$ (set to $|\mathcal{D}|-1$). \\
$\mathcal{B}_{d_i}$ & ID set that includes target token $d_i$ plus $|\mathcal{D}|-1$ reference IDs (used when probing preference for $d_i$). \\
$\pi_j^*(d_i)$ & Special ID permutation placing target token $d_i$ at position $j$, keeping others' relative order. \\
$\lambda_j^*$ & Special candidate permutation placing the matched resume at position $j$, keeping others' relative order. \\
$\pi^g$ & Randomly sampled ID permutation for job $g$ (used to approximately average out token bias in Eq. (\ref{eq13})). \\

\addlinespace
\multicolumn{2}{l}{\textbf{Bias analysis: probability quantities}}\\
$\mathbb{P}(d_i\mid \mathcal{B}_{d_i})$ & Preference/probability of selecting token $d_i$ given the constructed ID set $\mathcal{B}_{d_i}$ (conceptual definition). \\
$\tilde{\mathbb{P}}(d_i\mid \mathcal{B}_{d_i})$ & Approximated overall preference for $d_i$ obtained by averaging over positions under restricted permutations. \\
$\mathbb{P}(\text{position }j\mid \mathcal{D})$ & Preference for selecting the item at position $j$ when IDs are from set $\mathcal{D}$ (conceptual definition). \\
$\tilde{\mathbb{P}}(\text{position }j\mid \mathcal{D})$ & Approximated position preference using sampled $\pi_g$ and restricted $\lambda_j^*$. \\
$\pi(j)$ & Index (in the original ID set $\mathcal{D}$) of the ID token placed at position $j$ under permutation $\pi$. \\
$d_{\pi(j)}$ & The ID token appearing at position $j$ under permutation $\pi$. \\

\addlinespace
\multicolumn{2}{l}{\textbf{Debiasing derivations: observed vs.\ true preference}}\\
$f_{\lambda}(c_i)$ & Returns the position/index of candidate $c_i$ under candidate permutation $\lambda$. \\
$\mathbb{P}(c_i\mid g,\mathcal{D},\mathbf{C}_g)$ & ``True'' preference distribution over candidates for job $g$, marginalizing over permutations (conceptual). \\
$\mathbb{P}_{\mathrm{obs}}(c_i,d_j\mid g,\pi,\lambda,\mathcal{D},\mathbf{C}_g)$ & Observed joint distribution between candidate $c_i$ and token $d_j$ under a specific $(\pi,\lambda)$ (from the model's output). \\
$\mathbb{P}_{\mathrm{obs}}(d_j\mid g,\pi,\lambda,\mathcal{D},\mathbf{C}_g)$ & Observed probability of selecting token $d_j$ (typically the implicit first-token distribution). \\
$\mathbb{P}_{\mathrm{prior}}(\cdot)$ & Prior distribution over ID tokens capturing token/position prior bias in the decomposition assumption. \\
$\tilde{\pi}$ & Default permutation/order of ID tokens, i.e., the canonical order $\{\text{A}, \text{B}, \cdots\}$. \\
$\tilde{\lambda}$ & Default permutation/order of the candidate list, where resumes are arranged in alphabetical order. \\
\end{longtable}

\section{Theoretical Analysis of Local Positional Encoding}\label{appendix_a}
\begin{figure}[!htbp]
    \centering
    \includegraphics[width=0.6\linewidth]{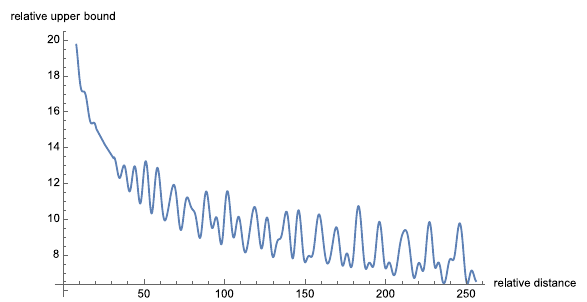}
    \caption{Long-term Decay of RoPE (from \citet{DBLP:journals/ijon/SuALPBL24}).}
    \label{figa1}
\end{figure}

To begin with, we briefly introduce the rotary positional encoding (RoPE)~\citep{DBLP:journals/ijon/SuALPBL24}. Throughout this section, we follow the notation used in the original RoPE paper for consistency. Given the query vector $\boldsymbol{q}_m \in \mathbb{R}^d$ at position $m$ and the key vector $\boldsymbol{k}_n \in \mathbb{R}^d$ at position $n$, RoPE applies a position-dependent rotation to each of them.  For example, the rotation matrix at position $m$ is
\begin{align*}
    \boldsymbol{\mathcal{R}_m} = {\left(\begin{array}{ccccccc}
  \cos m \theta_0 & -\sin m \theta_0 & 0 & 0 & \cdots & 0 & 0 \\
  \sin m \theta_0 & \cos m \theta_0 & 0 & 0 & \cdots & 0 & 0 \\
  0 & 0 & \cos m \theta_1 & -\sin m \theta_1 & \cdots & 0 & 0 \\
  0 & 0 & \sin m \theta_1 & \cos m \theta_1 & \cdots & 0 & 0 \\
  \vdots & \vdots & \vdots & \vdots & \ddots & \vdots & \vdots \\
  0 & 0 & 0 & 0 & \cdots & \cos m \theta_{d / 2-1} & -\sin m \theta_{d / 2-1} \\
  0 & 0 & 0 & 0 & \cdots & \sin m \theta_{d / 2-1} & \cos m \theta_{d / 2-1}
  \end{array}\right)}
\end{align*}
where $\boldsymbol{\mathcal{R}}_m$ is block diagonal, orthogonal, and satisfies $\boldsymbol{\mathcal { R }}_m\boldsymbol{\mathcal { R }}_m^{\top} = I$.  After applying $\boldsymbol{\mathcal{R}}_m$ and $\boldsymbol{\mathcal{R}}_n$ to $\boldsymbol{q}_m$ and $\boldsymbol{k}_n$, respectively, the inner product becomes
\begin{align*}
    (\boldsymbol{\mathcal { R }}_m \boldsymbol{q}_m)^{\top}(\boldsymbol{\mathcal { R }}_n \boldsymbol{k}_n)
= \boldsymbol{q}_m^{\top} \boldsymbol{\mathcal { R }}_{n-m} \boldsymbol{k}_n,
\end{align*}
where $\boldsymbol{\mathcal { R }}_{n-m}$ encodes \emph{relative position information} due to $\boldsymbol{\mathcal { R }}_m^{\top} \boldsymbol{\mathcal { R }}_n = \boldsymbol{\mathcal { R }}_{n-m}$.
RoPE exhibits \emph{long-term decay}: the value $\boldsymbol{q}^{\top} \boldsymbol{\mathcal { R }}_{n-m} \boldsymbol{k}$ decreases as the relative distance $|n-m|$ grows. The frequencies $\theta_i$ follow the sinusoidal scheme from the original Transformer~\citep{DBLP:conf/nips/VaswaniSPUJGKP17}, i.e., $\theta_i = 10000^{-2i/d}$. We can group the entries of $\boldsymbol{q}$ and $\boldsymbol{k}$ in pairs and use the complex notations to express the inner product as follows.
\begin{align}\label{eq20}
(\boldsymbol{\mathcal { R }}_m \boldsymbol{q})^{\top}(\boldsymbol{\mathcal { R }}_n \boldsymbol{k})
= \operatorname{Re}\left[\sum_{i=0}^{d / 2-1} \boldsymbol{q}_{[2 i: 2 i+1]} \boldsymbol{k}_{[2 i: 2 i+1]}^* e^{\mathrm{i}(m-n) \theta_i}\right].
\end{align}
where $\boldsymbol{q}_{[2i:2i+1]}$ is the $2i$-th to $2i+1$-th entries of $\boldsymbol{q}$. Let $h_i=\boldsymbol{q}_{[2 i: 2 i+1]} \boldsymbol{k}_{[2 i: 2 i+1]}^*$ and $S_j=\sum_{i=0}^{j-1} e^{\mathrm{i}(m-n) \theta_i}$, with $h_{d / 2}=0$ and $S_0=0$. By Abel's transform:
\[
 \sum_{i=0}^{d / 2-1} \boldsymbol{q}_{[2 i: 2 i+1]} \boldsymbol{k}_{[2 i: 2 i+1]}^* \mathrm{e}^{\mathrm{i}(m-n) \theta_i} = \sum_{i=0}^{d / 2-1} h_i (S_{i+1}-S_i)
= -\sum_{i=0}^{d / 2-1} S_{i+1}(h_{i+1}-h_i).
\]
This yields
$$
\begin{aligned}
\left|\sum_{i=0}^{d / 2-1} \boldsymbol{q}_{[2 i: 2 i+1]} \boldsymbol{k}_{[2 i: 2 i+1]}^* e^{\mathrm{i}(m-n) \theta_i}\right| & =\left|\sum_{i=0}^{d / 2-1} S_{i+1}\left(h_{i+1}-h_i\right)\right| \\
& \leq \sum_{i=0}^{d / 2-1}\left|S_{i+1}\right|\left|h_{i+1}-h_i\right| \\
& \leq\left(\max _i\left|h_{i+1}-h_i\right|\right) \sum_{i=0}^{d / 2-1}\left|S_{i+1}\right|
\end{aligned}
$$
so the decay behavior can be analyzed via the average $\frac{1}{d/2} \sum_{i=1}^{d/2} |S_i|$ as a function of $m-n$. Figure~\ref{figa1} shows the results reported by \citet{DBLP:journals/ijon/SuALPBL24}.

\begin{figure}[!htbp]
    \centering
    \includegraphics[width=0.6\linewidth]{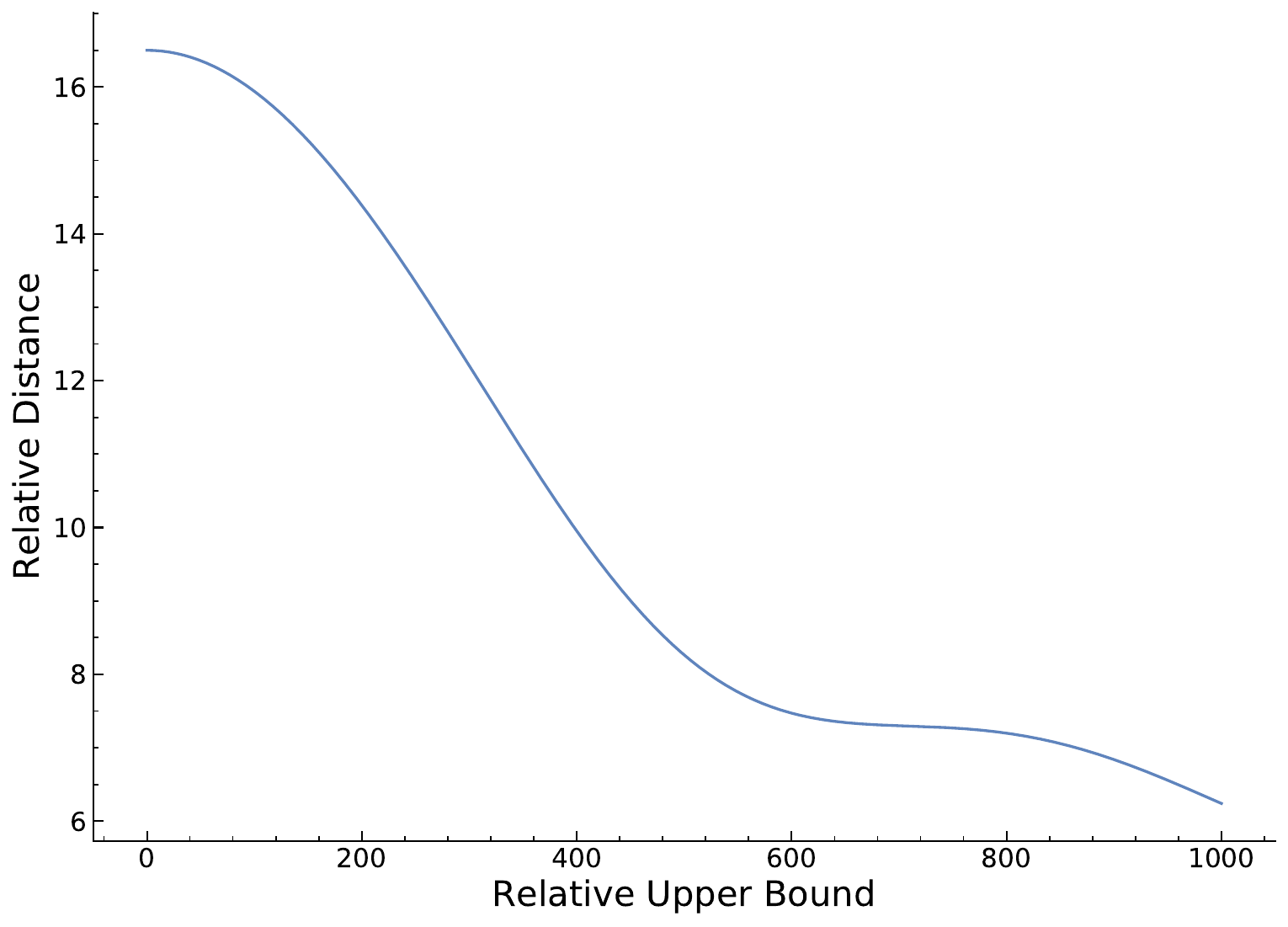}
    \caption{The Variation of $\frac{1}{d / 4} \sum^{d / 2}_{i=d/4+1} |S_i'|$ with Respect to Relative Distance $m'-n$ ($d = 128$).}
    \label{figa2}
\end{figure}

The prompt in listwise talent recommendation has a special structure, which begins with the task description and job posting, followed by each resume in the candidate set, as shown in Eq. (\ref{eq5}). Therefore, for the token that belongs to the resume $c_j$, we can calculate its local position as illustrated in Figure~\ref{fig3}. Now we consider a hierarchical modification of RoPE, which incorporates both global position $m$ and local position $m'$ into different subspaces of the rotation matrix. Let $\boldsymbol{\mathcal{R}}_m$ be decomposed into block diagonals
\[
\boldsymbol{\mathcal{R}}_m = \mathrm{diag}(\boldsymbol{\mathcal{R}}_{m,0}, \boldsymbol{\mathcal{R}}_{m,1}, \dots, \boldsymbol{\mathcal{R}}_{m,d/2-1}),
\]
where each $\boldsymbol{\mathcal{R}}_{m,i}$ is a $2\times 2$ matrix as follows.
$$
\boldsymbol{\mathcal{R}}_{m,i}  =  \left(\begin{array}{ccccccc}
  \cos m \theta_i & -\sin m \theta_i \\
  \sin m \theta_i & \cos m \theta_i  \\
  \end{array}\right)
$$
Suppose that the local position of the $m$-th token is $m'$ (e.g., within a resume), typically $m'\le m$, while a token at position $n$ belongs to a job posting with local position $n$. To inject local positional information, we replace a subset of $\boldsymbol{\mathcal{R}}_{m,i}$ with $\boldsymbol{\mathcal{R}}_{m',i}$:
\begin{align*}
  \boldsymbol{\mathcal{R}}_m' & = \mathrm{diag}{\left(\boldsymbol{\mathcal{R}}_{m,0}, \boldsymbol{\mathcal{R}}_{m,1}, ... \boldsymbol{\mathcal{R}}_{m,d/4-1}, \boldsymbol{\mathcal{R}}_{{\color{orange}m'},d/4},..., \boldsymbol{\mathcal{R}}_{{\color{orange}m'},d/2-1}\right)} \\
  \boldsymbol{\mathcal{R}}_{m,i} & =  \left(\begin{array}{ccccccc}
    \cos m \theta_i & -\sin m \theta_i \\
    \sin m \theta_i & \cos m \theta_i  \\
    \end{array}\right), 
  \boldsymbol{\mathcal{R}}_{{\color{orange}m'},i}  =  \left(\begin{array}{ccccccc}
      \cos {\color{orange}m'} \theta_i & -\sin {\color{orange}m'} \theta_i \\
      \sin {\color{orange}m'} \theta_i & \cos {\color{orange}m'} \theta_i  \\
      \end{array}\right)
\end{align*}
This changes the inner product to
\begin{align}\label{eq21}
  \left(\boldsymbol{\mathcal { R }}_m' \boldsymbol{q}\right)^{\top}\left(\boldsymbol{\mathcal { R }}_n \boldsymbol{k}\right) & =\operatorname{Re}\left[\underbrace{\sum_{i=0}^{d / 4 -1} \boldsymbol{q}_{[2 i: 2 i+1]} \boldsymbol{k}_{[2 i: 2 i+1]}^* e^{\mathrm{i}(m-n) \theta_i}}_{\text{Part 1}} + \underbrace{\sum_{i=d/4}^{d / 2-1} \boldsymbol{q}_{[2 i: 2 i+1]} \boldsymbol{k}_{[2 i: 2 i+1]}^* e^{\mathrm{i}({\color{orange}m'}-n) \theta_i}}_{\text{Part 2}}\right] 
\end{align}
For comparison, we can rewrite Eq. (\ref{eq20}) into two parts.
\begin{align}\label{eq22}
    \left(\boldsymbol{\mathcal { R }}_m \boldsymbol{q}\right)^{\top}\left(\boldsymbol{\mathcal { R }}_n \boldsymbol{k}\right) & =\operatorname{Re}\left[\underbrace{\sum_{i=0}^{d / 4 -1} \boldsymbol{q}_{[2 i: 2 i+1]} \boldsymbol{k}_{[2 i: 2 i+1]}^* e^{\mathrm{i}(m-n) \theta_i}}_{\text{Part 1}} + \underbrace{\sum_{i=d / 4}^{d/2-1} \boldsymbol{q}_{[2 i: 2 i+1]} \boldsymbol{k}_{[2 i: 2 i+1]}^* e^{\mathrm{i}({m}-n) \theta_i}}_{\text{Part 2}}\right] 
\end{align}
Eq. (\ref{eq21}) and Eq. (\ref{eq22}) share the same Part 1. Therefore, we focus on Part 2. Let $h_i = \boldsymbol{q}_{[2 i: 2 i+1]} \boldsymbol{k}_{[2 i: 2 i+1]}^*$ and $S_j' = \sum_{i=d/4}^{j-1} e^{\mathrm{i}({m'}-n) \theta_i}$, with $h_{d/2} = 0$ and $S_{d/4}' = 0$. By Abel's formula:
\begin{align*}
\sum_{i=d/4}^{d / 2-1} h_i\left(S_{i+1}'-S_i'\right)
&= - \sum_{i=d/4}^{d / 2-1} S_{i+1}'\left(h_{i+1}-h_i\right)
\end{align*}
and thus
$$
\left|\sum_{i=d/4}^{d / 2-1} h_i e^{\mathrm{i}({m'}-n) \theta_i}\right|
\leq \left(\max _i |h_{i+1}-h_i|\right) \sum_{i=d/4}^{d / 2-1} |S_{i+1}'|
$$
We examine $\frac{1}{d / 4} \sum^{d / 2}_{i=d/4+1} |S_i'|$ as a function of $m'-n$, as shown in Figure~\ref{figa2}.
From the figure, we conclude that Part 2 also exhibits long-distance decay. Replacing $m$ with $m'$ ($m' < m$) increases the value of Part 2, thereby increasing the overall dot product $\left(\boldsymbol{\mathcal { R }}_m' \boldsymbol{q}\right)^{\top}\left(\boldsymbol{\mathcal { R }}_n \boldsymbol{k}\right)$. In other words, injecting local position enhances the attention between resume tokens ($m$) and job posting tokens ($n$).

\begin{figure}[!htbp]
    \centering
    \includegraphics[width=0.6\linewidth]{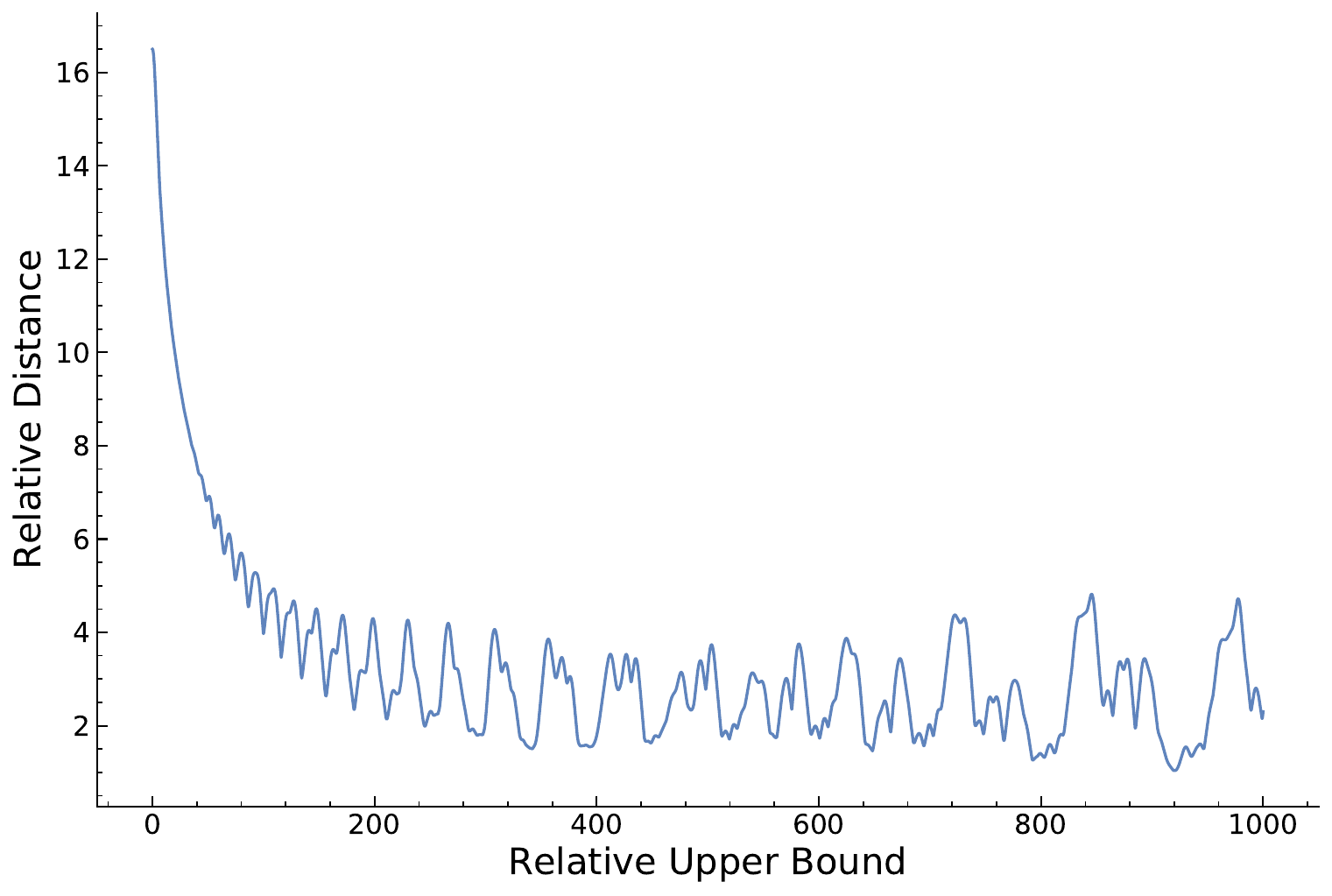}
    \caption{The Variation of $ \frac{1}{d / 2} \sum_{i=0}^{d / 2} \left| S_i \right| $ with Respect to Relative Distance $m'-n$ ($d = 128$).}
    \label{figa3}
\end{figure}

Next, consider using local position in Part 1 while retaining global position in Part 2:
\begin{align*}
\left(\boldsymbol{\mathcal { R }}_m' \boldsymbol{q}\right)^{\top}\left(\boldsymbol{\mathcal { R }}_n \boldsymbol{k}\right)
& =\operatorname{Re}\left[
\underbrace{\sum_{i=0}^{d / 4 -1} \boldsymbol{q}_{[2 i: 2 i+1]} \boldsymbol{k}_{[2 i: 2 i+1]}^* e^{\mathrm{i}({\color{orange}m'}-n) \theta_i}}_{\text{Part 1}} +
\underbrace{\sum_{i=d/4}^{d / 2-1} \boldsymbol{q}_{[2 i: 2 i+1]} \boldsymbol{k}_{[2 i: 2 i+1]}^* e^{\mathrm{i}(m-n) \theta_i}}_{\text{Part 2}}
\right]
\end{align*}
Define reversed indices:
\begin{align*}
h_i &= \boldsymbol{q}_{[2 (d/4-1-i): 2(d/4-1-i)+1]} \boldsymbol{k}_{[2 (d/4-1-i): 2 (d/4-1-i)+1]}^* \\
S_j' & = \sum_{i=0}^{j-1} e^{\mathrm{i}({\color{orange}m'}-n) \theta_{d/4-1-i}}
\end{align*}
with $h_{d/4} = 0$ and $S_0' = 0$. Applying Abel's formula again:
\begin{align*}
\sum_{i=0}^{d / 4-1} h_i\left(S_{i+1}'-S_i'\right)
&= - \sum_{i=0}^{d / 4-1} S_{i+1}'\left(h_{i+1}-h_i\right)
\end{align*}
Hence,
$$
\left|\sum_{i=0}^{d / 4-1} h_i e^{\mathrm{i}(m-n) \theta_i}\right|
\leq \left(\max _i |h_{i+1}-h_i|\right) \sum_{i=0}^{d / 4-1} |S_{i+1}'|
$$
We then examine $\frac{1}{d / 4} \sum^{d/4}_{i=1} |S_i'|$ versus $m'-n$, shown in Figure~\ref{figa3}.
We can summarize that replacing either the first half or the second half of $\boldsymbol{\mathcal{R}}_m$ with $\boldsymbol{\mathcal{R}}_{m',*}$ strengthens the attention between $m$ and $n$. Partitioning $\boldsymbol{\mathcal{R}}_m$ into three parts:
$$
\boldsymbol{\mathcal{R}}_m =
\mathrm{diag}{\left(
\underbrace{\boldsymbol{\mathcal{R}}_{m,0}, \dots, \boldsymbol{\mathcal{R}}_{m,a_1}}_{\text{Part 1}},
\underbrace{\boldsymbol{\mathcal{R}}_{m,a_1+1}, \dots, \boldsymbol{\mathcal{R}}_{m,a_2}}_{\text{Part 2}},
\underbrace{\boldsymbol{\mathcal{R}}_{m,a_2+1}, \dots, \boldsymbol{\mathcal{R}}_{m,d/2-1}}_{\text{Part 3}}
\right)}
$$
Compare:
\begin{align*}
\boldsymbol{\mathcal{R}}_m^{1} & =
\mathrm{diag}{\left(
\underbrace{\boldsymbol{\mathcal{R}}_{m,0}, \dots, \boldsymbol{\mathcal{R}}_{m,a_1}}_{\text{Part 1}},
\underbrace{\boldsymbol{\mathcal{R}}_{{\color{orange}m'},a_1+1}, \dots, \boldsymbol{\mathcal{R}}_{{\color{orange}m'},a_2}}_{\text{Part 2}},
\underbrace{\boldsymbol{\mathcal{R}}_{{\color{orange}m'},a_2+1}, \dots, \boldsymbol{\mathcal{R}}_{{\color{orange}m'},d/2-1}}_{\text{Part 3}}
\right)} \\
\boldsymbol{\mathcal{R}}_m^{2} & =
\mathrm{diag}{\left(
\underbrace{\boldsymbol{\mathcal{R}}_{m,0}, \dots, \boldsymbol{\mathcal{R}}_{m,a_1}}_{\text{Part 1}},
\underbrace{\boldsymbol{\mathcal{R}}_{{m},a_1+1}, \dots, \boldsymbol{\mathcal{R}}_{{m},a_2}}_{\text{Part 2}},
\underbrace{\boldsymbol{\mathcal{R}}_{{\color{orange}m'},a_2+1}, \dots, \boldsymbol{\mathcal{R}}_{{\color{orange}m'},d/2-1}}_{\text{Part 3}}
\right)}
\end{align*}
Considering only Part 2 and Part 3, the earlier analysis implies that $\boldsymbol{\mathcal{R}}_m^{1}$ will enhance the attention between $m$ and $n$ more than $\boldsymbol{\mathcal{R}}_m^{2}$. This means that the greater the dimensional proportion occupied by local position information, the stronger the attention between positions $m$ and $n$, with the ideal case being the complete adoption of local position information. Therefore, in L3TR, we replace the global position of each token by its corresponding local position. 

\section{Data}\label{appendix_b}
Two datasets are involved in our experiments. We provide more information about our data and the process for data preprocessing.

Table \ref{tab:app_a1} shows a pair of matched resumes and job postings from HRT. The original dataset is in Chinese, and the translated sample is presented in the table. HRT contains all information on job postings and anonymous information on resumes. During the experiments, we leverage the title, duty, and requirements in job postings and the educational background, job track, and project experience (if available) in resumes. Due to the limited space, we omit the details of each job in the job track. Note that some talents will provide important projects that they have led or worked on in their project experience. 
\begin{table}[!htbp]
\small
  \centering
    \caption{A Pair of Matched Resume and Job Posting from HRT.}
    \begin{tabular}{L{2cm}|L{5.5cm}|L{2cm}|L{5.5cm}}
    \toprule
    Job  & Description & Resume & Description \\
    \midrule
    Title  & Sales Manager & Education & Bachelor degree in Electronic Science and Technology\\
        \cline{2-2} \cline{4-4}
    Company & A company for enterprise employee training & Current Position & General Sales Manager   \\
    \cline{2-2} \cline{4-4} 
    Duty &  \underline{\it 1.} Develop business customer cooperation across the country through a variety of channels (telephone, network, exhibition, chamber of commerce, will sell, visit, friend recommendation, etc.); \underline{\it 2.} Mine customer needs and develop customers according to the company's product characteristics; \underline{\it 3.} Introduce and promote the company's mobile learning platform to enterprise customers; \underline{\it 4.} Design learning plans and customize new media courses for customers according to their needs; \underline{\it 5.} Complete the established customer development and sales targets on time; \underline{\it 6.} Complete the work assigned by the leader & \multirow{2}{*}{Job Track} &  \multirow{2}{5.5cm}{\underline{\it 1.} General sales manager of a company for mobile CRM software in Shenzhen from 2015 to 2019; \underline{\it 2.} Area channel director of an Internet marketing software company from 2012 to 2015; \underline{\it 3.} Co-founder of a technology company for website construction, searching engine optimization software, nesting bidding promotion from 2011 to 2012; \underline{\it 4.} Sales manager in a famous website providing efficient network promotion plans from 2009 to 2011; \underline{\it 5.} Sales representative in a well-known talent website from 2005 to 2009.} \\
    \cline{2-2} 
    Requirement & \underline{\it 1.} 3 years and above work experience, master the systematic sales method; \underline{\it 2.} With the experience and ability to sell complex solutions for large customers; \underline{\it 3.} Strong executive ability, love learning, with strong analysis and summary ability; \underline{\it 4.} Have a clear self-cognition and goal direction, can withstand greater work pressure; \underline{\it 5.} Traditional IT industry distribution experience is preferred; \underline{\it 6.} Can adapt to business trips and other emergencies  &  \\
    \cline{2-2} \cline{4-4}
    Location & Shenzhen, China & Workplace & Shenzhen, China \\
    \bottomrule
    \end{tabular}%
  \label{tab:app_a1}
\end{table}%

Table \ref{tab:app_a2} shows a pair of user profiles and job postings from JobRec. The original dataset is in English. JobRec contains complete details of job postings, while resumes only provide limited information, including some demographic attributes and job titles from work experience.
\begin{table}[!htbp]
\footnotesize
  \centering
    \caption{A Pair of Matched Resume and Job Posting from JobRec.}
     \begin{tabular}{L{2cm}|L{9cm}|L{2cm}|L{2cm}}
    \toprule
    Job  & Description & Resume & Description \\
    \midrule
    Title  & Automotive Service Advisor & Degree Type & High School \\
        \cline{2-2} \cline{4-4}
    \multirow{4}{*}{Description} & \multirow{4}{9cm}{Briggs Nissan in Lawrence Kansas has an immediate opportunity for the right person to join our team as a Service Advisor. Due to growth and expansion in our department we are looking for the right person that is looking for a career with our team.   Outstanding customer relation skills required. ADP computer experience a plus. Service department experience a plus but not required. Some Saturdays required. Excellent benefit package. If you feel you have what it takes be become part of our team.   Please contact,   Steve Schons, Service Manager Briggs Nissan, Lawrence Ks. (785) 856-7199} & Total Work Experience & 5.0 years   \\
    \cline{4-4} 
     & &  \\
      & & Currently Employed & Yes \\
      \cline{4-4} 
      & & \\
     & & In a Management Position & Yes\\
      \cline{4-4} 
      & & \\ 
     & & Number of People Managed & 10 \\
      \cline{4-4} 
    \cline{2-2} \cline{4-4} 
    Duty and Requirement &  The Service Advisor is responsible for building strong customer relationship and selling the technicians' time. They greet and consult customers on service needs, perform a thorough vehicle walk-around inspection as part of the write-up, sell and upsell services by emphasizing value, keep customers updated on services, field all live service calls, and take ownership of the customer's experience by carrying out those additional assignments that allow the dealership to leave an impressionable experience with the customer. Ensure that customers receive prompt, courteous, and effective service. Greet customers and assist them with any inquiries they may have. Perform vehicle walk around and documentation of customer concerns to enable the Technician to properly diagnose and service the vehicle. Drive the sale of technicians' time to meet the department's sales forecast. Consult with the customer on applicable service specials. Prioritize required services, and be prepared to provide options upon request. Document declines for services and ask for follow-up on future service considerations. Establish and maintain a close relationship with the customer. Keep the customer informed on completion times, service expenses, and possible changes. Remain involved with the delivery of the vehicle to the customer upon completion to ensure all customer concerns can be addressed. Ensure the customer has a positive dealership experience.  Provide concierge support for all owner inquiries, whether by phone or in person, to ensure the customer does not get mishandled. Spend quality time building a relationship with the customer. Know the product well enough to answer characteristic and operational questions regarding the customer's vehicle. Job Requirements: As a Service Advisor, you will be experienced and aware of the latest customer service practices, and be a persistent problem solver. We have determined some factors that may enable your success as a Service Advisor: One year of experience in a service advisor role is strongly preferred. Must have computer proficiency. Valid driver's license     & Work Experience & 1. Plane Captain on AH-1W, 2. Sales Representative, 3. Sales Representative and Technician \\
    \cline{2-2}  \cline{4-4} 
    Location & Lawrence, Kansas, US & Location &  Lenexa, Kansas, US\\
    \bottomrule
    \end{tabular}%
  \label{tab:app_a2}
\end{table}%

\section{Prompt Template}\label{appendix_c}
For HRT, the pointwise and listwise recommendation templates are presented in Table~\ref{table_c1} and Table~\ref{table_c2}, respectively. For explicit methods, we use the templates in the tables and parse the matching scores or the ranked list from the textual outputs. For implicit methods, we will remove the format instructions and calculate the output probability of special tokens to generate the final ranking. Specifically, for pointwise methods, the special tokens are ``yes'' and ``no''. While for listwise methods, the special tokens are the ID tokens. 
\begin{table}[!htbp]
  \centering
     \caption{Pointwise Recommendation Template for HRT.}   \label{table_c1}
  \begin{tabular}{L{1\linewidth}}
    \toprule
  \texttt{You are responsible for recruiting for a vacant position in a company. Please analyze whether the candidate is suitable for this position, based on the candidate's resume and the job posting. } \\
    \texttt{The job posting is as follows. $\{g\}$ } \\
\texttt{Resume $\{c_1^g\}$}\\
\texttt{Please output in a JSON format. The referenced output is as follows. } \\
\texttt{ \{  "score": the matching score of the candidate and the job, which should be between 0 and 1.\}} \\
    \bottomrule
  \end{tabular}
\end{table}

\begin{table}[!htbp]
  \centering
     \caption{Listwise Recommendation Template for HRT.}   \label{table_c2}
  \begin{tabular}{L{1\linewidth}}
    \toprule
  \texttt{You are responsible for recruiting for a vacant position in a company, please rank the resumes of {$\{N\}$} candidates based on the job description. The ranking should be determined by the matching degree between the resume and the job posting, with candidates who have a higher matching degree placed at the top. } \\
    \texttt{The job posting is as follows. $\{g\}$ } \\
\texttt{Here are the resumes for {$\{N\}$} candidates, each with a unique ID. Please use the IDs to sort all the resumes.
} \\
\texttt{Resume ID: A}\\
\texttt{Content: $\{c_1^g\}$} \\
\texttt{Resume ID: B}\\
\texttt{Content: $\{c_2^g\}$} \\
\texttt{...} \\
\texttt{Please directly output the sorted resume IDs without any additional content in JSON format. You should output in the following format. } \\
\texttt{ \{  "result": ["A", "B", ...] \}} \\
    \bottomrule
  \end{tabular}
\end{table}

For JobRec, the pointwise and listwise recommendation templates are presented in Table~\ref{table_c3} and Table~\ref{table_c4}, respectively. Since the notations in this paper are defined in the context of talent recommendation, they differ slightly from those used in job recommendations. Therefore, we do not include the notation in the tables. The templates follow roughly the same structure as those in HRT, except that the positions of the job posting and candidate information are swapped. Their usage in explicit and implicit methods remains largely consistent.
\begin{table}[t]
  \centering
     \caption{Pointwise Recommendation Template for JobRec.}   \label{table_c3}
  \begin{tabular}{L{1\linewidth}}
    \toprule
  \texttt{You are an expert at recruitment. Please analyze whether the candidate would be interested in this job based on the information of the candidate and the job posting. } \\
    \texttt{The detailed information of the candidate is as follows. $\{\}$ } \\
\texttt{The job posting is as follows. $\{\}$}\\
\texttt{Please output in a JSON format. The referenced output is as follows. } \\
\texttt{ \{  "score": the matching score of the candidate and the job, which should be between 0 and 1.\}} \\
    \bottomrule
  \end{tabular}
\end{table}

\begin{table}[!htbp]
  \centering
     \caption{Listwise Recommendation Template for JobRec.}   \label{table_c4}
  \begin{tabular}{L{1\linewidth}}
    \toprule
  \texttt{You are an expert at recruitment. Please rank the following $\{N\}$ job postings based on the information of the candidate. Please list job postings that align more closely with the candidate's interests at the front of the list.} \\
    \texttt{The detailed information of the candidate is as follows. $\{\}$ } \\
\texttt{The following are $\{N\}$ job postings, each with an ID, which is used to rank all jobs.
} \\
\texttt{Job Posting ID:  A}\\
\texttt{Content: $\{\}$} \\
\texttt{Job Posting ID:  B}\\
\texttt{Content: $\{\}$} \\
\texttt{...} \\
\texttt{Please directly output the IDs of ranked job postings in a JSON format. You should output in the following format. } \\
\texttt{ \{  "result": ["A", "B", ...] \}} \\
    \bottomrule
  \end{tabular}
\end{table}

\section{Experiment Details}\label{appendix_d}
\subsection{Metrics}
For a job posting $g$ and its candidate set $\mathcal{C}_g$, each model will produce the predicted scores $\hat{\mathbf{y}}_g$. All candidates in $\mathcal{C}_g$ are ranked in descending order of $\hat{\mathbf{y}}_g$, denoted by the permutation $\tau_g$. $\tau_g(l)$ returns the index of the candidate ranked at position $l$. Similarly, we can get the ranking permutation $\tau_g^*$ based on the ground-truth matching labels $\{y_{g,c_j}\mid c_j \in \mathcal{C}_g\}$. Then the three metrics are calculated as follows.
\begin{itemize}
    \item NDCG at cut-off $K$ (ND@$K$). 
    For each job posting $g$, we first compute the Discounted Cumulative Gain (DCG) at cut-off $K$ based on the ranked list $\tau_g$:
    \begin{align*}
    \mathrm{DCG}_g\text{@}K = \sum_{l=1}^{K} \frac{2^{y_{g,\tau_g(l)}} - 1}{\log_2(l+1)}
    \end{align*}
    The corresponding ideal DCG (IDCG) is obtained by sorting the candidates in $\mathcal{C}_g$ according to the ground-truth labels in descending order:
    \begin{align*}
    \mathrm{IDCG}_g\text{@}K = \sum_{l=1}^{K} \frac{2^{y_{g,\tau_g^{*}(l)}} - 1}{\log_2(l+1)}
    \end{align*}
    The ND@$K$ for $g$ is defined as
    \begin{align*}
    \mathrm{ND}_g\text{@}K = \frac{\mathrm{DCG}_g\text{@}K}{\mathrm{IDCG}_g\text{@}K}
    \end{align*}
    Finally, we report the average result across all test samples.
    \item Recall at cut-off $K$ (R@$K$).
    For each job posting $g$, R@$K$ is defined as 
    \begin{align*}
        \mathrm{R}_g\text{@}K = \frac{\sum_{l=1}^{K} y_{g,\tau_g(l)}}{\sum_{c_j\in\mathcal{C}_g} y_{g,c_j}}
    \end{align*}
    The final R@$K$ over the test set is obtained by averaging over all samples.
    \item Mean Reciprocal Rank (MRR). For each job posting $g$, let $\mathrm{rank}_g^+$ denote the rank position of the first relevant (positive) candidate in $\tau_g$:
    \begin{align*}
    \mathrm{rank}_g^+ = \min\{l \mid y_{g,\tau_g(l)} = 1\}
    \end{align*}
    The reciprocal rank for sample $g$ is defined as $1 / \mathrm{rank}_g^+$. MRR is computed as the mean reciprocal rank across the entire test set.
\end{itemize}

\section{Additional Experiment Results}\label{appendix_e}
\subsection{Evaluation Results with Candidate Sets of Different Lengths}
We evaluate each model with $N=10, 20, 30$, and the results are shown in Tables~\ref{table_e1}, \ref{table_e2}, \ref{table_e3}, \ref{table_e4}, \ref{table_e5}, and \ref{table_e6}.

As the candidate set size increases from $N=10$ to $N=30$, all models exhibit a clear and consistent performance degradation across ranking metrics. Despite this shared trend, L3TR consistently achieves the best performance at all candidate sizes.

\begin{table}[!htbp]
  \caption{Recommendation Accuracy on HRT with $N=10$.}
  \label{table_e1}
  \centering
    \begin{tabular}{l|l|rrrrr}
    \toprule
    Type & {Model} & ND@5  &  ND@10 &  R@1 & R@5 &  MRR \\
    \midrule
    \multirow{4}{3cm}{Domain-specific Models}
    & PJFNN   & .3512 & .4319 & .1613 & .5336 & .3421 \\
    & BPJFNN  & .5189 & .5754 & .2684 & .7521 & .4723 \\
    & APJFNN  & .3965 & .4956 & .1759 & .6148 & .3728 \\
    & TRBERT  & .6186 & .6713 & .4015 & .8052 & .5839 \\
    \midrule
    \multirow{8}{3cm}{LLM-based Models}
    & List-Imp        & .2196 & .3184 & .0857 & .3698 & .2389 \\
    & List-Exp        & .8169 & .8335 & .7018 & .8476 & .7964 \\
    & Point-Imp       & .5398 & .6121 & .3379 & .7184 & .5187 \\ 
    & Point-Exp       & .8036 & .8162 & .6437 & .8369 & .7675 \\
    & TALLRec         & .8617 & .8701 & .7239 & .8986 & .8282 \\
    & Rank1           & .8014 & .8153 & .6662 & .9068 & .7763 \\
    & ReasonRank (7B) & .6809 & .7134 & .5482 & .7946 & .6658 \\
    & ReasonRank (32B)& .8218 & .8362 & .7296 & .9037 & .8056 \\   
    \midrule 
    \multirow{1}{3cm}{Our Models}
    & L3TR & {\bf .8694} & {\bf .8821} & {\bf .8337} & {\bf .9035} & {\bf .8626} \\
    \bottomrule
    \end{tabular}%
\end{table}%
  
\begin{table}[!htbp]
  \caption{Recommendation Accuracy on HRT with $N=20$.}
  \label{table_e2}
  \centering
    \begin{tabular}{l|l|rrrrr}
    \toprule
   Type & {Model} & ND@5  &  ND@10 &  R@1 & R@5 &  MRR \\
    \midrule
    \multirow{4}{3cm}{Domain-specific Models}
    & PJFNN   & .3316 & .4119 & .1478 & .5112 & .3247 \\
    & BPJFNN  & .4959 & .5512 & .2483 & .7214 & .4489 \\
    & APJFNN  & .3749 & .4723 & .1589 & .5881 & .3526 \\
    & TRBERT  & .5917 & .6438 & .3784 & .7716 & .5534 \\
    \midrule
    \multirow{8}{3cm}{LLM-based Models}
    & List-Imp        & .1958 & .2876 & .0713 & .3349 & .2128 \\
    & List-Exp        & .7716 & .7924 & .6519 & .8047 & .7485 \\
    & Point-Imp       & .5026 & .5741 & .3082 & .6715 & .4794 \\ 
    & Point-Exp       & .7582 & .7726 & .5997 & .7856 & .7209 \\
    & TALLRec         & .8251 & .8364 & .6843 & .8619 & .7978 \\
    & Rank1           & .7604 & .7742 & .6268 & .8635 & .7336 \\
    & ReasonRank (7B) & .6457 & .6789 & .5098 & .7486 & .6284 \\
    & ReasonRank (32B)& .7823 & .7981 & .6895 & .8597 & .7662 \\   
    \midrule 
    \multirow{1}{3cm}{Our Models}
    & L3TR & {\bf .8327} & {\bf .8476} & {\bf .7829} & {\bf .8581} & {\bf .8334} \\
    \bottomrule
    \end{tabular}%
\end{table}%

\begin{table}[!htbp]
  \caption{Recommendation Accuracy on HRT with $N=30$.}
  \label{table_e3}
  \centering
    \begin{tabular}{l|l|rrrrr}
    \toprule
    Type & {Model} & ND@5  &  ND@10 &  R@1 & R@5 &  MRR \\
    \midrule
    \multirow{4}{3cm}{Domain-specific Models}
    & PJFNN   & .3116 & .3925 & .1353 & .4879 & .3058 \\
    & BPJFNN  & .4689 & .5247 & .2328 & .6894 & .4202 \\
    & APJFNN  & .3538 & .4519 & .1471 & .5606 & .3314 \\
    & TRBERT  & .5629 & .6158 & .3576 & .7362 & .5221 \\
    \midrule
    \multirow{8}{3cm}{LLM-based Models}
    & List-Imp        & .1698 & .2534 & .0601 & .3011 & .1849 \\
    & List-Exp        & .7234 & .7471 & .6017 & .7623 & .6992 \\
    & Point-Imp       & .4621 & .5346 & .2771 & .6206 & .4368 \\ 
    & Point-Exp       & .7089 & .7258 & .5502 & .7432 & .6723 \\
    & TALLRec         & .7854 & .7992 & .6469 & .8294 & .7461 \\
    & Rank1           & .7158 & .7311 & .5879 & .8263 & .6877 \\
    & ReasonRank (7B) & .6089 & .6441 & .4743 & .7106 & .5898 \\
    & ReasonRank (32B)& .7416 & .7604 & .6489 & .8237 & .7224 \\   
    \midrule 
    \multirow{1}{3cm}{Our Models}
    & L3TR & {\bf .7956} & {\bf .8138} & {\bf .7482} & {\bf .8226} & {\bf .7839} \\
    \bottomrule
    \end{tabular}%
\end{table}%

\begin{table}[!htbp]
      \caption{Recommendation Accuracy on JobRec with $N=10$.}
      \label{table_e4}
      \centering
    \begin{tabular}{l|l|rrrrr}
    \toprule
    Type & {Model} & ND@5  &  ND@10 &  R@1 & R@5 &  MRR \\
    \midrule
    \multirow{4}{3cm}{Domain-specific Models} 
    & PJFNN  & .2143 & .3208 & .0786 & .3819 & .2387 \\
    & BPJFNN & .2071 & .3134 & .0745 & .3698 & .2321 \\
    & TRBERT & .2024 & .3056 & .0739 & .3652 & .2294 \\
    \midrule
    \multirow{8}{3cm}{LLM-based Models} 
    & List-Imp  & .6184 & .6639 & .5037 & .7485 & .6073 \\
    & List-Exp  & .7812 & .7964 & .6521 & .9126 & .7489 \\
    & Point-Imp & .7368 & .7689 & .5914 & .8897 & .7058 \\ 
    & Point-Exp & .6027 & .6495 & .4052 & .8016 & .5639 \\
    & TALLRec   & .9156 & .9264 & .8813 & .9637 & .9185 \\
    & Rank1     & .6798 & .7184 & .4982 & .8715 & .6389 \\
    & ReasonRank (7B)  & .6662 & .7051 & .5267 & .8248 & .6417 \\
    & ReasonRank (32B) & .7196 & .7428 & .6249 & .8487 & .7075 \\ 
    \midrule 
    \multirow{1}{3cm}{Our Models} 
    & L3TR & {\bf .9348} & {\bf .9419} & {\bf .9032} & {\bf .9816} & {\bf .9364} \\
    \bottomrule
    \end{tabular}%
\end{table}%

\begin{table}[!htbp]
      \caption{Recommendation Accuracy on JobRec with $N=20$. }
      \label{table_e5}
      \centering
    \begin{tabular}{l|l|rrrrr}
    \toprule
   Type & {Model} & ND@5  &  ND@10 &  R@1 & R@5 &  MRR \\
    \midrule
    \multirow{3}{3cm}{Domain-specific Models} & PJFNN &  .2049 & .3094 & .0676 & .3474 & .2244 \\
    & BPJFNN & .1978 & .3014 & .0642  & .3355 & .2185  \\
    & TRBERT & .1937 & .2940 & .0636 & .3301 & .2157\\
    \midrule
     \multirow{8}{3cm}{LLM-based Models} & List-Imp  & .5880 & .6367 & .4435 & .7083 & .5755 \\
    & List-Exp  & .7426 & .7626 & .5754 & .8711 & .7104 \\
    & Point-Imp & .7012 & .7376 & .5171 & .8487 & .6668 \\ 
    & Point-Exp & .5713 & .6213 & .3482 & .7617 & .5319 \\
    & TALLRec & .8868 & .9016 & .8100 & .9341 & .8857 \\
    & Rank1 & .6461 & .6885 & .4316 & .8294 & .6031 \\
    & ReasonRank (7B) & .6333 & .6760 & .4556 & .7821 & .6049 \\
    & ReasonRank (32B) & .6831 & .7109 & .5515 & .8001 & .6691 \\ 
    \midrule 
    \multirow{1}{3cm}{Our Models} & L3TR & {\bf .9089} & {\bf .9201} & {\bf .8442} & {\bf .9578} & {\bf .9065} \\
    \bottomrule
    \end{tabular}%
\end{table}%

\begin{table}[!htbp]
      \caption{Recommendation Accuracy on JobRec with $N=30$. }
      \label{table_e6}
      \centering
    \begin{tabular}{l|l|rrrrr}
    \toprule
    Type & {Model} & ND@5  &  ND@10 &  R@1 & R@5 &  MRR \\
    \midrule
    \multirow{3}{3cm}{Domain-specific Models} & PJFNN &  .2006 & .3031 & .0638 & .3392 & .2187 \\
    & BPJFNN & .1939 & .2956 & .0604  & .3281 & .2129  \\
    & TRBERT & .1896 & .2887 & .0597 & .3215 & .2103\\
    \midrule
     \multirow{8}{3cm}{LLM-based Models} & List-Imp  & .5732 & .6234 & .4159 & .6861 & .5587 \\
    & List-Exp  & .7214 & .7438 & .5362 & .8475 & .6899 \\
    & Point-Imp & .6801 & .7176 & .4853 & .8224 & .6445 \\ 
    & Point-Exp & .5560 & .6052 & .3206 & .7368 & .5136 \\
    & TALLRec & .8617 & .8794 & .7682 & .9068 & .8584 \\
    & Rank1 & .6268 & .6681 & .4037 & .8032 & .5836 \\
    & ReasonRank (7B) & .6147 & .6579 & .4278 & .7576 & .5867 \\
    & ReasonRank (32B) & .6624 & .6908 & .5216 & .7745 & .6462 \\ 
    \midrule 
    \multirow{1}{3cm}{Our Models} & L3TR & {\bf .8836} & {\bf .8975} & {\bf .7931} & {\bf .9319} & {\bf .8812} \\
    \bottomrule
    \end{tabular}%
\end{table}%

\subsection{Evaluation Results with Training Sets of Different lengths}
We train L3TR with $N_{\text{train}}=3,5,6$ and evaluate models with $N=15$. The results are shown in Table \ref{table_e7}. 

Our results show that increasing $N_{\mathrm{train}}$ does not consistently improve the performance of L3TR$^{-}$. When the training prompt becomes longer, the model may struggle to extract meaningful ranking patterns, leading to marginal gains or even slight degradation.

In contrast, L3TR benefits from moderately larger $N_{\mathrm{train}}$, thanks to its local position encoding and specialized attention design, which enable it to better utilize structural information in longer prompts. However, the gains saturate beyond $N_{\mathrm{train}}=4$ or 5. Empirically, setting $N_{\mathrm{train}}$ to 4–5 is sufficient for handling ranking tasks with candidate list lengths of 15–30.

\begin{table}[!htbp]
      \caption{Recommendation Accuracy on HRT with Different $N_{\mathrm{train}}$. }
      \label{table_e7}
      \centering
    \begin{tabular}{l|l|rrrrr}
    \toprule
    Model & $N_{\mathrm{train}}$ & ND@5  &  ND@10 &  R@1 & R@5 &  MRR \\
    \midrule
    \multirow{4}{*}{L3TR$^-$ } &  3&  .7671 & .8010 & .6960 & .8290 & .7629 \\
    &  4 & .7692 & .7999 & .7046 & .8284 & .7682 \\
    & 5 & .7640 & .7965 & .6955 & .8230 & .7625 \\
    & 6 & .7585 & .7920 & .6900 & .8190 & .7570 \\
    \midrule
    \multirow{4}{*}{L3TR}  & 3 & .8235 & .8490 & .7780  & .8695 & .8250  \\
    & 4 &{.8379}& {.8545} & {.7953} & {.8762} & {.8377} \\
    & 5 & .8412 & .8595 & .8050 & .8795 & .8430 \\
    & 6 & .8482 & .8613 & .8074 & .8802 & .8451 \\
    \bottomrule
    \end{tabular}%
\end{table}%

\subsection{Evaluation Results with Different $S$}
We train L3TR with $S=0,1,5,15$ and evaluate models with $N=15$. The results in Table~\ref{table_e8} show that the number of special tokens $S$ plays an important role in enabling effective listwise interaction among resumes. When $S=0$, attention cannot be computed across different resumes. so even with listwise input, the model behaves similarly to pointwise methods (e.g., TALLRec in Table \ref{table3}). As $S$ increases, the model gains more capacity to facilitate information exchange across resumes through attention mechanisms. This enables the model to better compare candidates within the list, leading to consistent improvements in ranking performance. However, the marginal gain gradually diminishes when $S$ becomes sufficiently large, suggesting that a moderate number of special tokens (e.g., 10) is sufficient to support effective listwise recommendations. We recommend setting \(S\) to be sufficiently large to cover the resume identifier, at minimum including the following tokens: “[learnable token] Resume, ID: [Resume ID]”.

\begin{table}[!htbp]
      \caption{Recommendation Accuracy on HRT with Different $S$. }
      \label{table_e8}
      \centering
    \begin{tabular}{l|l|rrrrr}
    \toprule
    Model & $S$ & ND@5  &  ND@10 &  R@1 & R@5 &  MRR \\
    \midrule
    \multirow{5}{*}{L3TR}  & 0 & .8143 & .8366 & .7706  & .8548 & .8159  \\
    & 1 & .8185 & .8375 & .7706 & .8614 & .8211  \\
    & 5 & .8262 & .8473 & .7872  & .8630 & .8257  \\
    & 10 & {.8379}& {.8545} & {.7953} & {.8762} & {.8377} \\
    & 15 & .8291 & .8512 & .7859 & .8688 & .8310 \\
    \bottomrule
    \end{tabular}%
\end{table}%

\subsection{Additional Bias Evaluation Results}
We present the estimated prior bias (i.e., $\hat{\mathbb{P}}(d_i\mid \mathcal{D})$ in Eq. (\ref{eq18})) of L3TR, L3TR$^{-}$, and List-Imp (c.r. Table~\ref{table7}) in Table~\ref{table_e9}. The results show that List-Imp exhibits severe prior bias, with probability mass heavily concentrated on ID token ``A'', as reflected by the extremely large range and standard deviation. In contrast, L3TR produces a much more uniform prior distribution, achieving the smallest range and standard deviation.

\begin{table}[!htbp]
      \caption{Estimated Prior Bias on HRT with $N=15$.}
      \label{table_e9}
      \centering
\resizebox{0.99\linewidth}{!}{
    \begin{tabular}{c|rrrrrr}
    \toprule
    ID Token & List-Imp & List-Imp+Pre-ranker & L3TR$^-$  & L3TR$^-$+Pre-ranker &  L3TR & L3TR+Pre-ranker  \\
    \midrule
    ``A'' & 0.4739 & 0.5168 & 0.0826 & 0.3291 & 0.0563 & 0.3133 \\
    ``B'' & 0.0424 & 0.0378 & 0.0915 & 0.1278 & 0.0842 & 0.1357 \\
    ``C'' & 0.0309 & 0.0318 & 0.1451 & 0.1396 & 0.0770 & 0.1011 \\
    ``D'' & 0.2786 & 0.2426 & 0.0867 & 0.0681 & 0.0636 & 0.0602 \\
    ``E'' & 0.0204 & 0.0201 & 0.1120 & 0.0667 & 0.0829 & 0.0579 \\
    ``F'' & 0.0243 & 0.0267 & 0.0241 & 0.0150 & 0.0654 & 0.0514 \\
    ``G'' & 0.0129 & 0.0112 & 0.0618 & 0.0565 & 0.0621 & 0.0582 \\
    ``H'' & 0.0029 & 0.0036 & 0.0750 & 0.0501 & 0.0728 & 0.0420 \\
    ``I'' & 0.0004 & 0.0004 & 0.0473 & 0.0283 & 0.0608 & 0.0364 \\
    ``J'' & 0.0047 & 0.0064 & 0.0256 & 0.0080 & 0.0634 & 0.0206 \\
    ``K'' & 0.0014 & 0.0017 & 0.0632 & 0.0326 & 0.0853 & 0.0303 \\
    ``L'' & 0.0615 & 0.0642 & 0.0517 & 0.0352 & 0.0532 & 0.0372 \\
    ``M'' & 0.0059 & 0.0063 & 0.0380 & 0.0183 & 0.0580 & 0.0312 \\
    ``N'' & 0.0189 & 0.0139 & 0.0183 & 0.0078 & 0.0392 & 0.0123 \\
    ``O'' & 0.0211 & 0.0155 & 0.0770 & 0.0168 & 0.0756 & 0.0121 \\
    \midrule
Range & 0.4735 & 0.5164 & 0.1268 & 0.3213 & 0.0290 & 0.3012 \\
Std. Dev. & 0.1277 & 0.1414 & 0.0374 & 0.0794 & 0.0100 & 0.0755 \\
    \bottomrule
    \end{tabular}%
    }
\end{table}%

\end{document}